\documentclass[10pt,twocolumn,letterpaper]{article}

\usepackage{cvpr}      
\usepackage{multirow}

\usepackage[utf8]{inputenc} 
\usepackage[T1]{fontenc}    
\usepackage{url}            
\usepackage{booktabs}       
\usepackage{amsfonts}       
\usepackage{nicefrac}       
\usepackage{microtype}      
\usepackage{xcolor}         
\usepackage{amsmath}
\usepackage{graphicx} 
\usepackage{amssymb}
\usepackage{booktabs}
\usepackage{color}
\usepackage{amsmath,bm}
\usepackage{multirow}
\usepackage{float}
\usepackage{subcaption} 
\usepackage{tabularray}
\usepackage{pifont}
\usepackage{enumitem}
\usepackage{colortbl}
\usepackage{algorithm}
\usepackage{makecell}
\usepackage{float} 
\definecolor{cvprblue}{rgb}{0.21,0.49,0.74}
\usepackage[pagebackref,breaklinks,colorlinks,allcolors=cvprblue]{hyperref}

\def\paperID{2666} 
\def\confName{CVPR}
\def\confYear{2025}

\title{\textcolor{black}{Multi-Sensor Object Anomaly Detection: \\Unifying  Appearance, Geometry, and Internal Properties}}

\author{
    $\text{Wenqiao Li}^{1}\thanks{Equal contribution}$ \quad
    $\text{Bozhong Zheng}^{1\ast}$ \quad
    $\text{Xiaohao Xu}^{2\ast}$ \quad
    $\text{Jinye Gan}^{1}$ \quad 
    $\text{Fading Lu}^{1}$ \quad \\
    $\text{Xiang Li}^{1}$ \quad
    $\text{Na Ni}^{1}$ \quad 
    $\text{Zheng Tian}^{1}$  \quad 
    $\text{Xiaonan Huang}^{2}$\quad
    $\text{Shenghua Gao}^{3\dagger}$ \quad
    $\text{Yingna Wu}^{1}\thanks{Corresponding authors}$ \\
    $^{1}$ShanghaiTech University \quad $^{2}$University of Michigan, Ann Arbor \quad $^{3}$The University of Hong Kong \\}
\begin{document}
\maketitle

\begin{abstract}
\textcolor{black}{Object anomaly detection is essential for industrial quality inspection, yet traditional single-sensor methods face critical limitations. They fail to capture the wide range of anomaly types, as single sensors are often constrained to either external appearance, geometric structure, or internal properties. To overcome these challenges, we introduce MulSen-AD, the first high-resolution, multi-sensor anomaly detection dataset tailored for industrial applications. MulSen-AD unifies data from RGB cameras, laser scanners, and lock-in infrared thermography, effectively capturing external appearance, geometric deformations, and internal defects. The dataset spans 15 industrial products with diverse, real-world anomalies. We also present MulSen-AD Bench, a benchmark designed to evaluate multi-sensor methods, and propose MulSen-TripleAD, a decision-level fusion algorithm that integrates these three modalities for robust, unsupervised object anomaly detection. Our experiments demonstrate that multi-sensor fusion substantially outperforms single-sensor approaches, achieving 96.1\% AUROC in object-level detection accuracy. These results highlight the importance of integrating multi-sensor data for comprehensive industrial anomaly detection. The dataset and code are available at \url{https://github.com/ZZZBBBZZZ/MulSen-AD} to support further research.}%
\end{abstract}    
\begin{figure}[ht]
  \centering
   \includegraphics[width=1.0\linewidth]{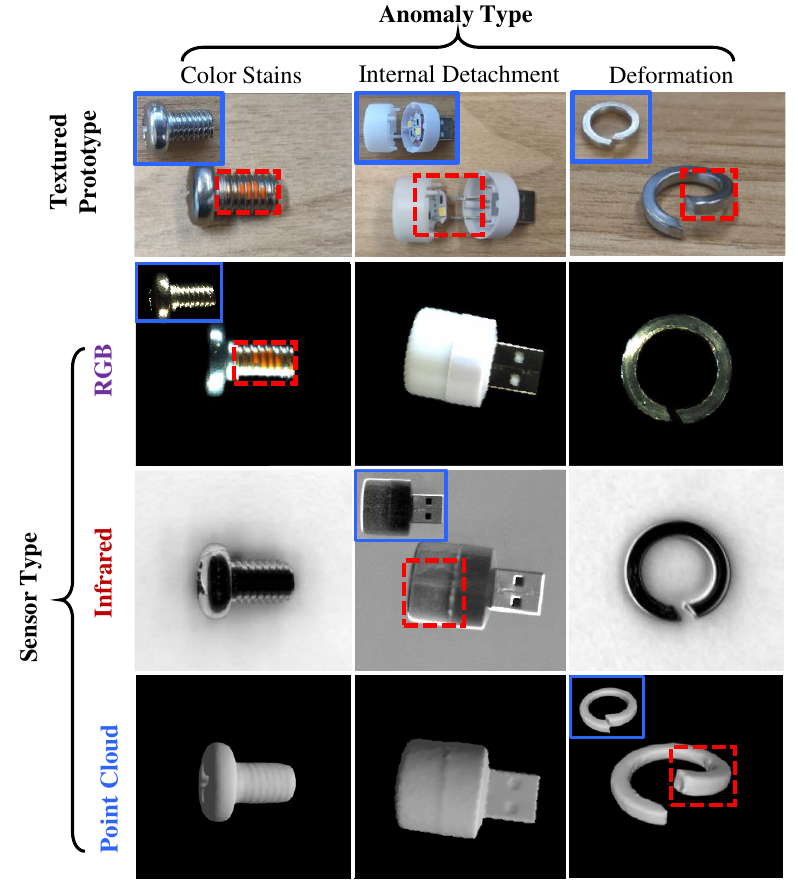}
      \caption{\textcolor{black}{
\textbf{Motivation for multi-sensor object anomaly detection.}
Different sensors capture distinct anomalies, making fusion essential. Our  \textbf{\textit{MulSen-AD}} dataset demonstrates how RGB captures surface defects, point clouds detect geometric deformations, and infrared reveals internal and subsurface issues. Red boxes enclose anomalies, blue highlights normal references.}}
  \label{fig:1}\vspace{-2mm}
\end{figure}

\section{Introduction}
\label{sec:intro}

\textcolor{black}{
In the industrial manufacturing landscape, ensuring the quality and reliability of products is not just a matter of economic efficiency but also of consumer safety and brand reputation. {Object-level Anomaly Detection} plays a pivotal role in this context, serving as the first line of defense against defective products entering the market~\cite{cao2024survey,liu2024deep}. Despite significant advances~\cite{bergmann2022beyond,8954181,Mishra_2021,liu2024real3d,zou2022spotthedifference,CAO2024110761},  single-sensor anomaly detection methods face inherent limitations that hinder their effectiveness in real-world applications.}

\textcolor{black}{
\textbf{One of the fundamental challenges with single-sensor systems is their inability to capture the multifaceted nature of anomalies present in industrial products. }Anomalies can manifest in various forms, \textit{e.g.}, surface scratches, internal cracks, thermal inconsistencies, and structural deformations, that no single sensor modality can comprehensively detect. For instance, while an RGB camera excels at capturing color and texture variations on the surface, it cannot detect subsurface defects. Conversely, a laser scanner might capture geometric distortions but miss thermal anomalies indicative of internal stresses. This limitation is illustrated in Figure~\ref{fig:1}, where we demonstrate the complementary strengths of each sensor modality in our MulSen-AD dataset. RGB sensors focus on external appearance defects, infrared sensors detect subsurface and internal anomalies, and point cloud sensors identify 3D geometric deformations. As shown in the teaser figure, multi-sensor fusion is essential to achieve robust and comprehensive anomaly detection by leveraging the strengths of each sensor.}


\textcolor{black}{
To address these gaps, we introduce MulSen-AD, the first multi-sensor anomaly detection dataset designed specifically for industrial applications. This dataset integrates high-resolution RGB images, infrared thermal images, and high-precision 3D point cloud data from laser scanners, offering a comprehensive resource for detecting a wide variety of real-world defects. Unlike existing datasets, MulSen-AD captures anomalies that span multiple modalities, providing a richer evaluation framework for anomaly detection models.}

\textcolor{black}{
In addition to the dataset, we also propose a baseline model, MulSen-TripleAD, which leverages multi-sensor fusion for anomaly detection. This baseline combines data from RGB, infrared, and point cloud sensors, using decision-level fusion to achieve more accurate and robust anomaly detection. Our experiments demonstrate that MulSen-TripleAD outperforms single-sensor models, achieving an AUROC of 96.1\%, significantly higher than the results obtained from models relying on a single sensor. These findings underscore the critical importance of multi-sensor data fusion in capturing a broader range of anomalies and improving detection performance in industrial environments.
}

\textcolor{black}{
{Our main contributions are summarized as follows:}}
\begin{itemize}
    \item \textbf{MulSen-AD framework.} We introduce the \textit{MulSen-AD framework}, a novel multi-sensor approach for industrial object anomaly detection that integrates high-resolution RGB imaging, high-precision laser scanning, and lock-in infrared thermography to capture a comprehensive representation of anomalies.
    
    \item \textbf{MulSen-AD dataset.} We present \textit{MulSen-AD}, the first real-world dataset specifically designed for evaluating multi-sensor anomaly detection in industrial settings. The dataset features diverse, high-quality data from 15 distinct industrial products with real-world defects.
    
    \item \textbf{Benchmark and toolkit.} We establish a comprehensive benchmark on the MulSen-AD dataset and provide an open-source toolkit to support further research, ensuring ease of experimentation and reproducibility.
    
    \item \textbf{MulSen-TripleAD model.} We propose \textit{MulSen-TripleAD}, a decision-level fusion gating method for unsupervised multi-sensor anomaly detection. By combining data from the three sensor types and utilizing multiple memory banks with a decision gating unit, MulSen-TripleAD significantly outperforms single-sensor setups, achieving 96.1\% AUROC in object-level anomaly detection accuracy, which highlights importance of multi-sensor data.

\end{itemize}

\begin{table}[t!]
\centering
\caption{\textcolor{black}{Comparison of our MulSen-AD dataset with existing object-level anomaly detection datasets. `Syn', `IR', `D', and `PC' denote Synthetic, Infrared, Depth, and Point Cloud, respectively.} }
\setlength{\tabcolsep}{1pt}
\renewcommand\arraystretch{1.2}
\resizebox{\linewidth}{!}{
\begin{tabular}{@{}l|ccccccc@{}}
\toprule
\multirow{2}{*}{\textbf{Dataset}} &
  \multirow{2}{*}{\textbf{Year}} &
  \multirow{2}{*}{\textbf{Type}} &
  \multirow{2}{*}{\textbf{Modality}} &
  \multicolumn{3}{c}{\textbf{Sample Statistics}} \\ 
  \cmidrule(l){5-7} 
  
                    &      &        &            &Class & Anomaly Types   & Multi-Sensor                  \\ \midrule
{MVTec-AD}~\cite{8954181}       & 2019    & Real     & RGB     & 15     & -    & \ding{55} \\
{BTAD}~\cite{Mishra_2021}       & 2021   & Real     & RGB     & 3    & 3    & \ding{55} \\
{MPDD}~\cite{9631567}       & 2021   & Real     & RGB     & 6   &  8   & \ding{55} \\
VisA~\cite{zou2022spotthedifference}         & 2021   & Real     & RGB     & 12    & -   &    \ding{55}\\  
MVTec LOCO-AD~\cite{bergmann2022beyond}         & 2022   & Real     & RGB     & 5    & -   &    \ding{55}\\  

MAD~\cite{zhou2023pad}         & 2023   & Syn+Real     & RGB     & 20    & 3   &    \ding{55}\\ 
LOCO-Annotations~\cite{10710633}    & 2024   & Real     & RGB     & 5    & 5   &    \ding{55}\\  
Real-IAD~\cite{wang2024realiad}        & 2024          & Real     & RGB     & 30      & 8   & \ding{55} \\\midrule

GDXray~\cite{mery2015gdxray}         & 2015   & Real     & X-ray      & 5    & 15   &    \ding{55}\\  
PVEL-AD~\cite{9744494}         & 2023   & Real     & IR     & 1    & 10   &    \ding{55}\\  \midrule

MVTec3D-AD~\cite{Bergmann_2022}        & 2021         & Real     & RGB-D    & 10     & 3-5   & \ding{55} \\
Eyecandies~\cite{bonfiglioli2022eyecandies}        & 2022        & Syn    & RGB-D    & 10     & 3   & \ding{55} \\ \midrule

Real3D-AD~\cite{liu2024real3d}        & 2023        & Real    & PC   & 12    & 2   & \ding{55} \\
Anomaly-ShapeNet~\cite{li2023towards}        & 2024        & Syn   & PC   & 40    & 6  & \ding{55} \\ \midrule

\textbf{MulSen-AD (Ours)} & 2024  & Real & RGB\&IR\&PC     & 15     &14 & \ding{51}  \\ \bottomrule
\end{tabular}}
\label{tab:1}
\end{table}

\begin{figure*}[t!]
  \centering
  \setlength{\abovecaptionskip}{0.cm}
  \includegraphics[width=1.0\textwidth]{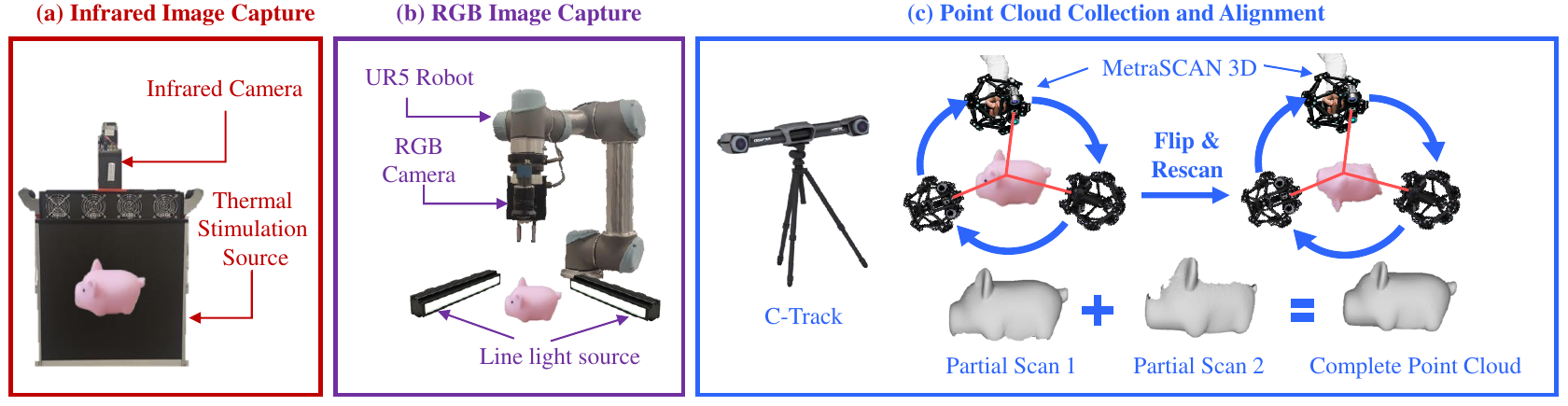}
  \caption{\textcolor{black}{\textbf{Data collection pipeline for the proposed MulSen-AD dataset} consists of three stages: (\textbf{a}) Infrared image acquisition, (\textbf{b}) RGB image capture, and (\textbf{c}) Point cloud collection and alignment. The pink 'Piggy' object serves as the example for data collection.}}
  \label{fig:device}\vspace{-2mm}
\end{figure*}

\section{Related work}\label{sec:related_work}
\textbf{Object-level anomaly detection datasets.}
Object-level anomaly detection aims to identify defective samples during or after industrial production processes. Historically, this field has relied solely on data from single sensor.  MVTec-AD~\cite{8954181}, BTAD~\cite{Mishra_2021}, MPDD~\cite{9631567}, and VisA~\cite{zou2022spotthedifference} is a series of single view photo-realistic industrial anomaly detection datasets. The objects provided in these datasets are just captured in one single view RGB camera. The overall shape information of objects cannot be captured, and texture information is easily affected by lighting and environmental conditions in this setting. The LOCO AD dataset~\cite{bergmann2022beyond} provides rich global structural and logical information but is not suitable for fine-grained anomaly detection on individual objects, which is extended by LOCO-Annotations~\cite{10710633}. MVTec3D-AD~\cite{Bergmann_2022} and Eyecandies~\cite{bonfiglioli2022eyecandies} intend to integrate depth maps with RGB maps to provide the geometry information under the fixing single view. MAD~\cite{zhou2023pad} and Real-IAD~\cite{wang2024realiad} are multi-view AD datasets, trying to provide texture information and depth information from different views. Visual anomaly detection under RGB cameras cannot avoid being easily influenced by ambient lighting and the confusion in detecting superficial morphological abnormalities. As a solution for this, datasets such as PVEL-AD~\cite{9744494} and GDXray~\cite{mery2015gdxray} intend to detect external and shallow layer anomalies within an object by infrared and X-ray sensors. Nonetheless, PVEL-AD and GDXray sacrifice color and texture information.  Real3D-AD~\cite{liu2024real3d} and Anomaly-ShapeNet~\cite{li2023towards} are 3D AD datasets, which only focus on object-level geometry anomaly detection. In a word, existing object AD datasets just rely on one single kind of sensor, which often fail to accurately capture all types of anomalies in actual factory settings, significantly limiting the advancement of this area. To address these challenges and explore the problem of multi-sensor anomaly detection, we propose MulSen-AD, the first dataset that includes RGB images, infrared images, and high-resolution point clouds specifically for anomaly detection. As shown in Table~\ref{tab:1}, MulSen-AD uniquely covers all three modalities—RGB images, infrared images, and high-quality point cloud data—setting it apart from existing object anomaly detection datasets.

\begin{table}[t]
  \centering
    \caption{\textbf{\textcolor{black}{Data collection device parameters.}}}
    \setlength{\tabcolsep}{4pt}
    \renewcommand\arraystretch{1.1} 
    \resizebox{\linewidth}{!}{
    \begin{tabular}{l|c|c|c}
    \toprule
    \multirow{1}{*}{\textbf{Device}} & \begin{tabular}[c]{@{}c@{}}Daheng \\ MER2-230-168U3C\end{tabular}
      &  \begin{tabular}[c]{@{}c@{}}Noverlteq \\ TWILIS-180\end{tabular}   & \begin{tabular}[c]{@{}c@{}}Creaform \\ MetraSCAN 750\end{tabular}     \\ \midrule
      \textbf{Modality}  & RGB Image & Infrared Image & Point Cloud \\
    
    \textbf{Resolution} & \(1920 \times 1200\) & \(640 \times 480\) & 0.05 mm \\
    \textbf{Accuracy} & ---  & \(\pm 2^\circ\text{C}\) & 0.03 mm \\
 \textbf{Pixel Depth} & 8bit & 16bit & ---  \\
    \textbf{Wavelength Range} & ---  & \(7.5\text{--}14\,\mu\text{m}\) & ---  \\
    \textbf{Scanning Area} & ---  & ---  & \(275 \times 250\,\text{mm}\) \\
    \bottomrule
    \end{tabular}} \label{tab:2}
    \end{table}

\noindent\textbf{Multi-sensor fusion methods.}
Existing multi-sensor fusion methods can be categorized into data (early) fusion, feature (middle) fusion, and decision (late) fusion~\cite{alaba2024emerging}. Data fusion methods, \textit{e.g.}, PointPainting~\cite{Vora_2020_CVPR}, PointAugmenting~\cite{Wang_2021_CVPR}, MVP~\cite{yin2021multimodal}, and RVF-Net~\cite{nobis2021radar} aim to integrate data from various sources or sensors early, creating a unified representation that can be directly utilized for subsequent processing steps. Feature fusion methods such as DeepFusion~\cite{li2022deepfusion}, TransFusion~\cite{bai2022transfusion}, EPNet~\cite{huang2020epnet}, AutoAlignV2~\cite{chen2022deformable}, NS-MAE~\cite{xu2024self}, and DeepInteration~\cite{yang2022deepinteraction} strive to facilitate the transformation of input data into more abstract feature representations across different layers in the training phase, empowering the model to effectively utilize data at each network layers. Late fusion methods~\cite{pang2020clocs,zhang2021rangelvdet}, follow a strategy where data from multi-sensor is processed independently before being combined at the fusion stage, aiming to minimize errors caused by discrepancies in the data. Inspired by these methods, we adopt a late fusion strategy for multi-sensor anomaly detection, leveraging the strengths of independent sensor modalities before combining them at the decision-making stage to enhance robustness and mitigate sensor-specific errors.

%

\section{Dataset: MulSen-AD}
\label{construction}

\subsection{Sensor Selection}

\begin{figure*}[t!]
  \centering
  \setlength{\abovecaptionskip}{0.cm}
  \includegraphics[width=1.0\textwidth]{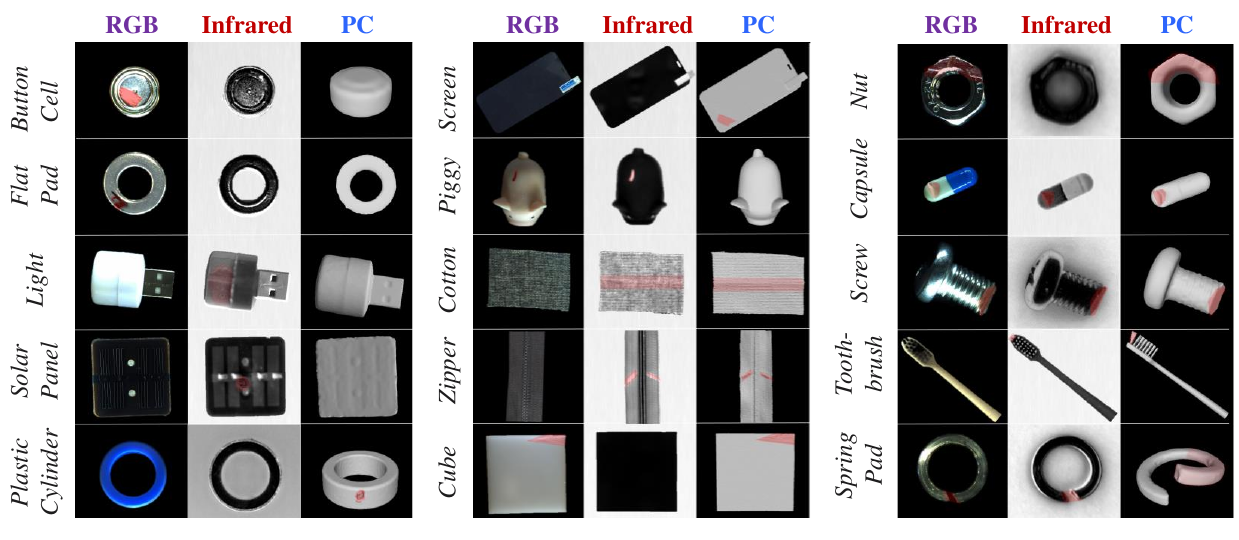}
  \caption{\textcolor{black}{{\textbf{15 object categories from MulSen-AD}, each represented in three modalities—RGB, IR, and point cloud. Some defects are visible in only one or two modalities. We highlight the abnormal areas using red overlay masks. }}}
  \label{fig:3}\vspace{-2mm}
\end{figure*}

\textcolor{black}{The most common sensors used in industrial applications include industrial cameras, infrared sensors, ultrasonic sensors, and 3D laser scanners. However, due to the complexity of interpreting ultrasonic signals—particularly the need to eliminate various types of noise—ultrasonic sensors were not included in our dataset. Thus, MulSen-AD incorporates RGB images from cameras, gray-scale images from lock-in infrared thermography, and high-resolution 3D point clouds from laser scanners.}

\textcolor{black}{
As shown in Figure~\ref{fig:device}, we illustrate the data collection pipeline for the multi-sensor dataset construction.
Detailed hyperparameters of the sensors used are listed in Table~\ref{tab:2}. These resources offer a comprehensive overview of the sensor configurations and the anomalies targeted in our dataset, ensuring clarity and consistency throughout the data collection and annotation processes.}


%

\noindent\textcolor{black}{\textbf{Lock-in infrared thermography.}
 Noverlteq TWILIS-180 lock-in infrared system with FLIR A600 infrared camera (640$\times$480 resolution) was applied to capture the IR gray-scale images. To detect anomalies, periodic thermal stimulation is applied to heat the objects. If the heat absorption of anomalies is different from the objects, the temperature difference is presented in the images. In our dataset, infrared camera successfully captures the temperature anomaly of broken inside capsule, damages in solar panels, detachment of parts inside lamps and the other internal anomalies that RGB camera cannot detect.}


\noindent\textcolor{black}{\textbf{RGB camera.} The Daheng MER2-230-168U3C camera, with a maximum resolution of 1920$\times$1200, was mounted on a UR5 robotic arm to capture images above the target object. Line light sources were positioned on either side of the object to ensure uniform lighting, minimizing shadows and enhancing the visibility of intricate surface details.} 

\noindent\textcolor{black}{\textbf{3D Scanner System.} The Creaform MetraSCAN 750 with HandyPROBE CMM captures high-precision point clouds using a hand-held scanner and C-Track sensor. Laser grids project onto objects, and reflected light is used to reconstruct 3D surfaces while tracking global coordinates. With 0.03 mm precision and 0.05 mm resolution, it captures 20k-100k points per object. The portable design allows 360-degree scanning, eliminating blind spots found in fixed systems like Zivid One-Plus (applied by MVTec3D-AD) and PMAX-S130 (applied by Real3D-AD), enabling detection of 3D anomalies such as spring pad deformations and cotton creases, which RGB and infrared cameras often miss.}

    


\subsection{Object Preparation}
\textcolor{black}{We selected 15 objects of varying materials (metal, plastic, fiber, rubber, semiconductor, composites) with diverse shapes, sizes, and colors. To replicate industrial conditions, we manually introduced 14 types of anomalies, including cracks, holes, breaks, creases, scratches, foreign bodies, labeling errors, bends, color defects, and detachments (Figure~\ref{fig:3}). These cover surface, internal, and 3D geometric defects, ensuring comprehensive representation of industrial anomaly scenarios.}

\subsection{Data Collection and Processing} 

\textcolor{black}{The data collection process involved capturing images from multiple sensors, as outlined in Figure~\ref{fig:3}.}

\noindent\textcolor{black}{\textbf{Infrared camera.} The infrared camera, positioned above the centrally placed object at random horizontal angles, captured grayscale images at a resolution of 640$\times$480. Objects were periodically heated using thermal stimulation sources for 30 to 180 seconds, depending on their material and thickness.}

\noindent\textcolor{black}{\textbf{RGB camera.} The RGB camera, with a resolution of 1280$\times$960, captured top-view images. Based on the infrared images, the position of the objects and the height of the camera were adjusted using a measuring scale, a UR5 robotic arm, and camera software grid.}

\noindent\textcolor{black}{\textbf{Point cloud.} 3D point clouds were acquired using a laser scanner. To ensure complete coverage, objects were flipped during the scanning process (\textit{dual-scan strategy}). The point clouds from each scan were manually coarsely aligned, followed by fine alignment using the Iterative Closest Point (ICP) algorithm. This process was repeated until the point clouds were accurately aligned. }


\subsection{Data Annotation}
\textcolor{black}{After collecting the data, we annotate pixel-level masks for anomalies in both RGB and infrared images using the \textit{LabelMe} tool. For point cloud data, we employ \textit{Geomagic Design X} to manually select the anomalous regions, saving the selected points in a text file format. Importantly, a modality will only receive annotations if the anomaly is visible in that specific modality. For instance, if an internal anomaly in a capsule is detected in the infrared image but is not visible in the RGB image or point cloud, only the infrared image is annotated. This modality-specific approach ensures precise and relevant labeling across all sensor types.}


\subsection{Data Statistics}
\noindent\textcolor{black}{\textbf{Dataset sample distribution.} Table~\ref{tab:3} presents the statistical information of MulSen-AD dataset, which includes the dataset category, the number of training set samples, the number of normal and abnormal samples in the test set, and the number of anomaly types. The MulSen-AD dataset comprises 2035 samples evenly distributed across 15 categories.  On average, 33 abnormal samples are included in the test set for each category, providing a diverse evaluation suite.}

\begin{table}[t!]
\centering
\caption{\textcolor{black}{\textbf{MulSen-AD dataset statistics.} The proportions of abnormal pixels and points are calculated exclusively for the abnormal samples. `PC': Point Cloud. The abnormal ratio is calculated either for pixel or point for each category.}}
\centering
\setlength{\tabcolsep}{0.5mm}
    \renewcommand\arraystretch{1}
\resizebox{0.48\textwidth}{!}
{
\begin{tabular}{clcccccccccc}
\toprule
& \multirow{2}{*}{\textbf{Category}} 
& \textbf{Train Set} 
& \multicolumn{2}{c}{\textbf{Test Set}} 
& \multirow{2}{*}{\#\textbf{Total}} 
& {\#\textbf{Anomaly}} 
& \multicolumn{3}{c}{\textbf{Abnormal  Ratio[\%]}} \\ 

\cmidrule(r){3-3} \cmidrule(r){4-5}  \cmidrule(r){8-10}  
         & &  \#\textit{Normal} & \#\textit{Normal} &\#\textit{Abnormal} & &   {\textbf{Classes}} &  RGB& Infrared&PC &  \\\midrule
 &Capsule & 64                  & 10         & 48            & 122   &  6   & 0.392 \  & 0.346\  & 11.1\ &   \\
  &Cotton & 78                 & 10         & 40            & 128   & 5  &  0.771 \ & 0.569\    & 3.32\ &            \\
&Cube  & 110                 & 10           & 41           & 161   & 5   &  0.558 \ & 0.552\    & 2.07\ &           \\
 & Piggy  &110                 & 10           & 30            & 150   & 5   & 0.444 \  & 0.444\    & 1.37\ &           \\
 &Screen & 69                  & 10          & 32            & 111   &  4  &  0.774 \ & 1.070\    & 4.28\ &           \\
  &Flat Pad  & 90                  & 10           & 30            & 130  & 4  & 0.188 \  & 0.193\    & 4.99\ &            \\
 & Screw  &90                 & 10           & 31           & 131   & 5  & 0.314 \  & 0.330\    &3.48\  &            \\
&Nut & 118                  & 10           & 29           & 157   & 4      & 0.201 \   & 0.117\ &  5.85\   &       \\
&Spring Pad & 86            & 10           & 24            & 120   &  5     & 0.056 \   & 0.078\ & 19.1\  &         \\
 &Button Cell  & 90           & 10          & 31            & 131   &  4     & 0.259 \   &0.227\  & 1.69\    &       \\
 &Toothbrush & 110        & 10          & 25           & 145   & 5     & 0.105 \   & 0.126\ & 7.58\  &\\
&Zipper & 86                 & 10          & 30            & 126   &  5     & 0.687 \   & 0.997\ & 6.19\   &        \\
&Light & 110         & 10          & 36           & 156   & 6     & 0.209\   & 0.838\ & 1.24\  &\\
&Plastic Cylinder & 90         & 10          & 28            & 128   & 5     & 0.317 \   & 0.427\ & 1.94\  &\\  
&Solar Panel & 90         & 10          & 39            & 139   & 5     & 0.306 \   & 0.458\ & 0.496\  &\\\midrule
&  Mean  &  93                  & 10          & 33          & 136   & 4.8 &  0.372 \  & 0.451\     & 4.98\  &  \\ 
& Total  & \textbf{1391}                & \textbf{150}        & \textbf{494}          & \textbf{2035}  & \textbf{72} & ---   & ---  &---  &  \\ \bottomrule
\end{tabular}}
\label{tab:3}
\end{table}

\begin{figure}[t!]
\centering
 \setlength{\abovecaptionskip}{0.cm}
\includegraphics[width=0.7\linewidth]{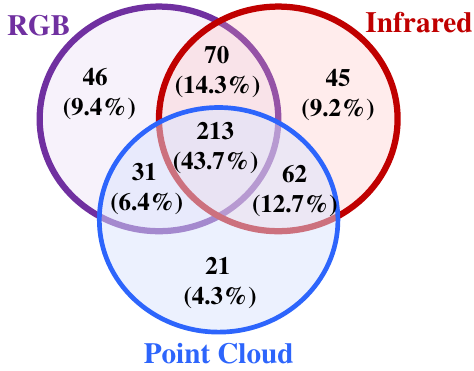}
    \captionof{figure}{\textcolor{black}{\textbf{Anomaly distribution captured by single and multiple sensors in MulSen-AD. } Overlap regions represent anomalies that are observable  by multiple sensors.}}
    \label{fig:vanne}
\end{figure}

\begin{figure*}[t]
  \centering
  \setlength{\abovecaptionskip}{0.cm}
  \includegraphics[width=1.0\textwidth]{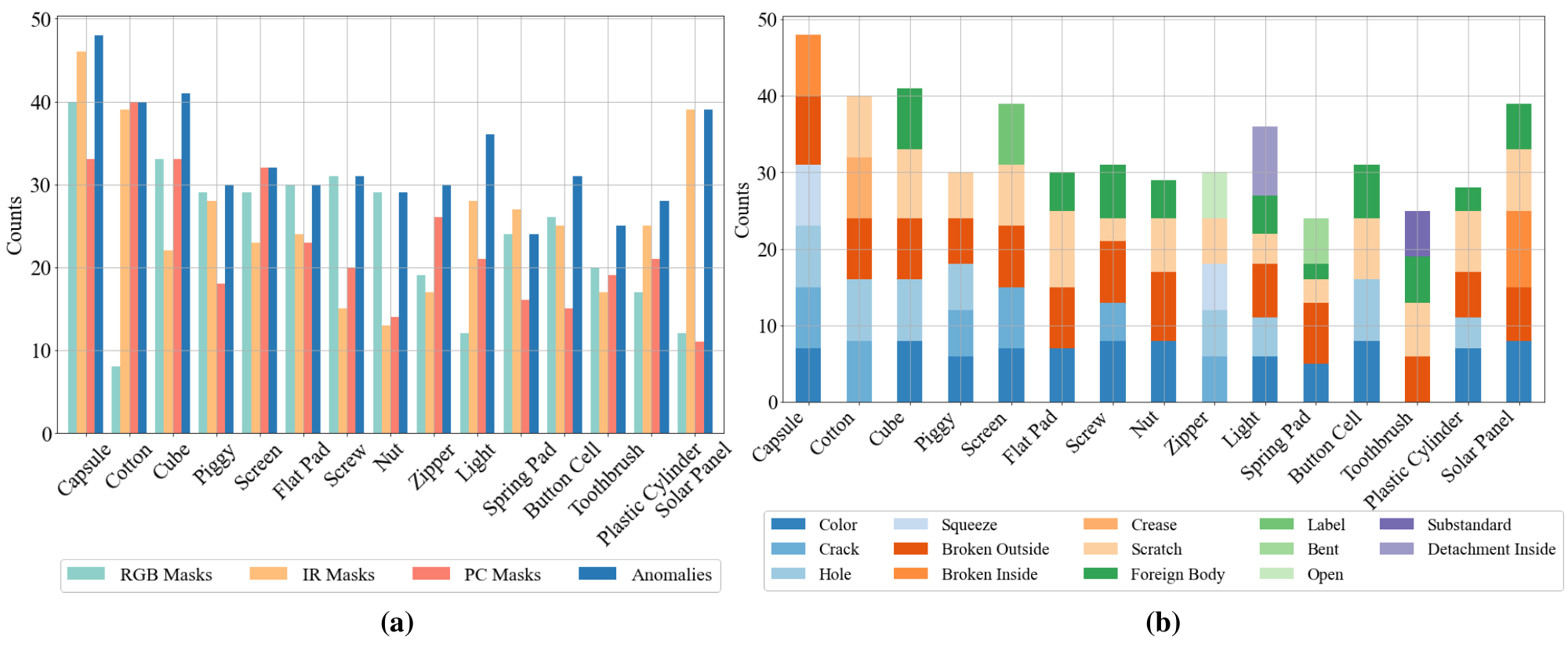}
  \caption{\textcolor{black}{\textbf{Anomaly data distribution of MulSen-AD dataset.} The annotation count for each modality reflects the number of detectable anomaly samples per modality. (\textbf{a}) Anomaly annotation counts by modality across categories. (\textbf{b}) Distribution of anomaly types per  category.
}}
  \label{fig:4}\vspace{-6mm}
\end{figure*}

\noindent\textcolor{black}{\textbf{Multi-sensor data complementarity.}
The Venn diagram in Figure~\ref{fig:vanne} demonstrates the distribution of anomalies detected by the RGB, infrared, and point cloud sensors in the MulSen-AD dataset. Non-overlapping regions highlight each sensor's ability to capture specific anomalies independently, such as the 9.4\% of anomalies detected solely by the RGB sensor, 9.2\% by infrared, and 4.3\% by point cloud. The overlapping areas indicate anomalies detected by multiple sensors, with 43.7\% of anomalies being identified by all three modalities.  }
\noindent\textcolor{black}{\textbf{Anomaly annotation distribution. } As shown in Figure~\ref{fig:4}-(a), it is obvious that the three modalities show different advantages in detecting anomalies in different categories. For instance, anomalies on solar panels are most effectively identified in infrared images, while anomalies on cotton are more accurately detected using point clouds. Figure~\ref{fig:4}-(b) displays the types of anomalies associated with each category, with an average of 4.8 defect types per category. }

\section{Baseline Model: MulSen-TripleAD}
\label{Baseline}
\subsection{Problem Definition}
\label{problem definition}

\textcolor{black}{
In this work, we focus on the \textbf{unsupervised anomaly detection setting for multi-sensor inputs}. The objective of the developed model is to accurately predict the object-level anomaly label $L_{\rm o}$ based on the multi-sensor inputs (RGB, infrared, and point cloud) in a zero-shot setting, where no labeled anomalies are seen during training. For this problem, we consider RGB, infrared, and point cloud sensors as the data input. The training set $\mathcal{T} = \left\{ t_i \right\}_{i=1}^{N}$ consists of anomaly-free objects, each represented by the three modalities. During testing, the model encounters objects from the same categories but with potential anomalies, without prior exposure to anomalous examples.}

\textcolor{black}{At test time, each sample comprises an RGB image $I_{\rm rgb}$, infrared image $I_{\rm ir}$, and 3D point cloud $P$. The anomaly detection task is to predict an object-level anomaly label $L_{\rm o}$, where an object is considered anomalous if at least one of the modality-specific labels is positive ($L=1$). If all modality labels are negative ($L=0$), the object is deemed normal.}


\begin{figure}
  \centering
 \setlength{\abovecaptionskip}{0.cm}
  \includegraphics[width=0.48\textwidth]{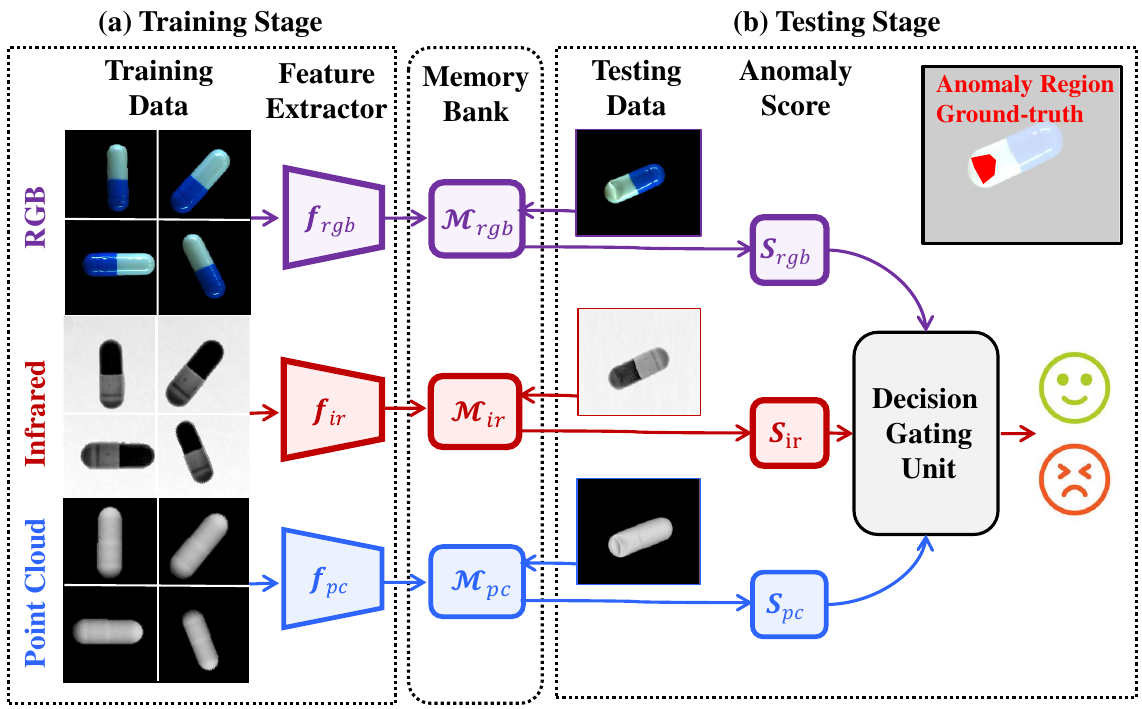}
  \caption{\textcolor{black}{\textbf{Pipeline overview of  MulSen-TripleAD method.}} }
  \label{fig:5}
\end{figure}%

\subsection{MulSen-TripleAD Method}
\label{sec:MulSen-TripleAD}

\textcolor{black}{Inspired by PatchCore~\cite{roth2022towards} and M3DM~\cite{wang2023multimodal}, we propose MulSen-TripleAD, a multi-sensor anomaly detection baseline, as illustrated in Figure~\ref{fig:5}. The MulSen-TripleAD framework is composed of three key components:}

\noindent\textcolor{black}{\textbf{Multi-modal feature extraction.} We utilize two pretrained feature extractors—DINO~\cite{zhang2022dino} for RGB and infrared images, and PointMAE~\cite{pang2022masked} for point clouds. These extractors generate distinct representations for each modality.}

\noindent\textcolor{black}{\textbf{Multi-modal memory bank establishment.} For each sensor, we construct a memory bank: $\mathcal{M}_{rgb}$ for RGB, $\mathcal{M}_{ir}$ for infrared, and $\mathcal{M}_{pc}$ for point clouds. These memory banks are built using normal samples during training, following the approach in PatchCore~\cite{roth2022towards}. During inference, each sensor’s memory bank is used to compute an anomaly score based on the deviation of the test sample from the normal data.}

\noindent\textcolor{black}{\textbf{Decision gating unit}.
After obtaining the anomaly scores from each sensor, we integrate these scores using a Decision Gating Unit $\mathcal G_a$, inspired by the learnable One-Class Support Vector Machine (OCSVM) from M3DM~\cite{wang2023multimodal}. The final object-level anomaly score, $S$, is calculated as: \begin{equation} S = \mathcal G_a (\phi(\mathcal{M}_{rgb},f_{rgb}), \phi(\mathcal{M}_{pt},f_{pt}), \phi(\mathcal{M}_{ir},f_{ir})), \label{eq
} \end{equation} where $\phi$ is the scoring function from PatchCore~\cite{roth2022towards}: \begin{equation}
     \phi(\mathcal{M}, f) = \Vert f^{(i,j)^*} - m^*\Vert_2,
\end{equation}
\begin{equation}
     f^{(i,j)^*}, m^* = \arg \max _{f^{(i,j)}\in f} \arg \min _{m\in \mathcal{M}}\Vert  f^{(i,j)} - m\Vert_2,
\end{equation}
where $\mathcal{M}$ refers to the memory bank for each modality ($\mathcal{M}_{rgb}$, $\mathcal{M}_{pt}$, or $\mathcal{M}_{ir}$), and $f$ represents the extracted features for each sensor. }

More implementation details of MulSen-TripleAD are provided in the supplementary material.


\begin{table*}[t!]
\centering
\caption{\textcolor{black}{\textbf{Anomaly detection results, measured by object-level AUROC ($\uparrow$).}} The best result in each category is highlighted in \textbf{bold}.}
\vspace{-7pt}
\centering\setlength{\tabcolsep}{3mm}
\resizebox{\textwidth}{!}{
\begin{tabular}{l|ccccccccccccccc|c}
 \toprule
\textbf{Method} & \textbf{Capsule} & \textbf{Cotton} & \textbf{Cube} & \makecell{\textbf{Spring} \\ \textbf{Pad}} & \textbf{Screw} & \textbf{Screen} & \textbf{Piggy} & \textbf{Nut} & \makecell{\textbf{Flat} \\ \textbf{Pad}} & \makecell{\textbf{Plastic} \\ \textbf{Cylinder}} & \textbf{Zipper} & \makecell{\textbf{Button} \\ \textbf{Cell}} & \makecell{\textbf{Tooth} \\ \textbf{brush}}  & \makecell{\textbf{Solar} \\ \textbf{Panel}} & \textbf{Light} & \textbf{Mean} \\ 
\midrule
\multicolumn{17}{c}{\textbf{RGB-based Anomaly Detection Methods}} \\ \midrule
CFA\cite{Lee2022CFA} & 0.865 & 0.979 & 0.875 & 0.822 & 0.981 & 0.302 & 0.978 & 0.814 & 0.370 & 0.891 & 0.617 & 0.731 & 0.593 & 0.777 & 0.841 & 0.762 \\
CFLOW-AD\cite{CFLOW} & 0.938 & \textbf{1.000} & 0.919 & 0.731 & 0.926 & 0.371 & 0.984 & 0.941 & 0.953 & 0.930 & 0.794 & 0.818 & 0.910 & 0.854 & 0.848 & 0.861 \\
DeSTSeg\cite{DeSTSeg} & 0.296 & 0.551 & 0.350 & 0.680 & 0.300 & 0.920 & 0.799 & 0.279 & 0.400 & 0.305 & 0.444 & 0.572 & 0.487 & 0.610 & 0.618 & 0.507 \\
DRAEM\cite{DRAEM} & 0.279 & 0.622 & 0.483 & 0.443 & 0.519 & 0.318 & 0.759 & 0.676 & 0.373 & 0.487 & 0.576 & 0.436 & 0.360 & 0.473 & 0.564 & 0.491 \\
InvAD\cite{InvAD} & 0.940 & 0.994 & 0.939 & 0.980 & 0.965 & 0.334 & 0.987 & 0.966 & 0.947 & 0.978 & 0.865 & 0.810 & 0.917 & 0.910 & 0.843 & 0.892 \\
PatchCore\cite{patchcore} & 0.778 & 0.619 & 0.861 & 0.955 & \textbf{1.000} & 0.884 & 0.955 & \textbf{1.000} & \textbf{1.000} & 0.738 & 0.786 & 0.833 & 0.867 & 0.635 & 0.647 & 0.837 \\
RD++\cite{rd++} & 0.804 & 0.997 & 0.939 & 0.901 & 0.939 & 0.379 & 0.940 & 0.966 & 0.907 & 0.958 & 0.805 & 0.818 & 0.947 & 0.890 & 0.745 & 0.862 \\
SimpleNet\cite{simplenet} & 0.906 & 0.994 & 0.897 & 0.885 & 0.955 & 0.565 & {0.994} & 0.855 & 0.897 & 0.930 & 0.812 & 0.803 & 0.817 & 0.755 & 0.728 & 0.853 \\
\midrule
\multicolumn{17}{c}{\textbf{Infrared-based Anomaly Detection Methods}} \\ \midrule
CFA\cite{Lee2022CFA} & 0.628 & 0.354 & 0.555 & 0.671 & 0.687 & 0.506 & 0.580 & 0.660 & 0.607 & 0.840 & 0.667 & 0.368 & 0.350 & 0.810 & 0.472 & 0.584 \\
CFLOW-AD\cite{CFLOW} & 0.858 & 0.909 & \textbf{0.987} & 0.913 & 0.628 & 0.572 & 0.958 & 0.521 & 0.969 & {0.997} & 0.893 & 0.800 & 0.788 & 0.595 & 0.797 & 0.812 \\
DeSTSeg\cite{DeSTSeg} & 0.541 & 0.655 & 0.511 & 0.498 & 0.533 & 0.341 & 0.586 & 0.391 & 0.510 & 0.508 & 0.771 & 0.490 & 0.690 & 0.597 & 0.468 & 0.539 \\
DRAEM\cite{DRAEM} & 0.577 & 0.890 & 0.586 & 0.567 & 0.441 & 0.432 & 0.551 & 0.710 & 0.422 & 0.323 & 0.586 & 0.518 & 0.660 & 0.426 & 0.663 & 0.557 \\
InvAD\cite{InvAD} & {0.960} & 0.900 & 0.986 & 0.882 & 0.677 & 0.307 & 0.976 & 0.760 & 0.958 & {0.988} & \textbf{0.997} & 0.818 & 0.784 & 0.662 & 0.825 & 0.832 \\
PatchCore\cite{patchcore} & 0.914 & 0.995 & \textbf{0.987} & 0.917 & 0.644 & 0.375 & 0.985 & 0.760 & 0.893 & \textbf{1.000} & 0.976 & 0.718 & 0.820 & 0.867 & 0.791 & 0.843 \\
RD++\cite{rd++} & 0.887 & 0.895 & 0.975 & 0.907 & 0.664 & 0.348 & 0.976 & 0.754 & 0.878 & 0.994 & 0.982 & 0.655 & 0.804 & 0.736 & 0.762 & 0.814 \\
SimpleNet\cite{simplenet} & 0.881 & 0.974 & 0.956 & 0.893 & 0.621 & 0.561 & 0.964 & 0.716 & 0.898 & 0.985 & {0.991} & 0.678 & 0.775 & 0.823 & 0.733 & 0.830 \\
\midrule
\multicolumn{17}{c}{\textbf{Point Cloud-based Anomaly Detection Methods}} \\ \midrule
BTF(FPFH)\cite{BTF} & 0.923 & 0.320 & 0.634 & 0.512 & 0.592 & 0.788 & 0.377 & 0.531 & 0.660 & 0.579 & 0.637 & 0.655 & 0.659 & 0.423 & 0.378 & 0.578 \\
BTF(Raw)\cite{BTF} & 0.829 & 0.775 & 0.447 & 0.383 & 0.908 & 0.584 & 0.360 & 0.459 & 0.373 & 0.404 & 0.479 & 0.645 & {0.924} & 0.308 & 0.442 & 0.555 \\
M3DM(PointMAE)\cite{m3dm} & 0.835 & 0.435 & 0.615 & 0.808 & 0.629 & 0.494 & 0.667 & 0.590 & 0.797 & 0.675 & 0.744 & 0.697 & 0.803 & 0.695 & 0.756 & 0.683 \\
M3DM(PointBERT)\cite{m3dm} & 0.604 & 0.548 & 0.192 & 0.308 & 0.787 & \textbf{0.953} & 0.167 & 0.586 & {0.910} & 0.264 & 0.649 & 0.571 & 0.890 & 0.541 & 0.428 & 0.560 \\
PatchCore(FPFH)\cite{patchcore} & 0.898 & 0.228 & 0.759 & 0.763 & 0.742 & 0.900 & 0.830 & 0.714 & 0.850 & 0.771 & {0.917} & 0.884 & {0.917} & 0.818 & 0.764 & 0.784 \\
PatchCore(FPFH+raw)\cite{patchcore} & 0.892 & 0.395 & 0.664 & 0.754 & 0.739 & {0.922} & 0.760 & 0.831 & 0.743 & 0.843 & 0.872 & 0.710 & 0.932 & 0.759 & 0.683 & 0.767 \\
PatchCore(PointMAE)\cite{patchcore} & 0.835 & 0.435 & 0.615 & 0.808 & 0.629 & 0.494 & 0.667 & 0.590 & 0.797 & 0.675 & 0.744 & 0.697 & 0.803 & 0.695 & 0.756 & 0.683 \\
Reg3D-AD\cite{real3d-ad} & 0.867 & 0.592 & 0.451 & 0.804 & 0.716 & 0.444 & 0.610 & 0.683 & 0.700 & 0.725 & 0.810 & 0.613 & 0.898 & 0.695 & 0.789 & 0.693 \\
\midrule
\textbf{MulSen-TripleAD (Ours)} & \textbf{0.967} & 0.960 & 0.980 & 0.879 & \textbf{1.000} & 0.938 & \textbf{1.000} & 0.959 & 0.863 & 0.993 & 0.994 & \textbf{1.000} & \textbf{0.955} & \textbf{0.949} & \textbf{0.972} & \textbf{0.961} \\
\bottomrule
\end{tabular}}
\label{tab:results}\vspace{-4mm}
\end{table*}

\begin{table}[t!]
\centering
\caption{\textcolor{black}{\textbf{Anomaly detection results, measured by object-level AUROC ($\uparrow$), of MulSen-TripleAD with different modality input on MulSen-AD. }`PC' refers to point cloud. `IR' refers to infrared image. The best result in each category is highlighted in \textbf{bold}.}}
\vspace{-7pt}
\centering\setlength{\tabcolsep}{2mm}
\resizebox{0.48\textwidth}{!}{
\begin{tabular}{lccccccc}
\\ \toprule
\multirow{2}{*}{\textbf{Category}} & \multicolumn{3}{c}{\textbf{Single}} & \multicolumn{3}{c}{\textbf{Dual}}  &  \multicolumn{1}{c}{\textbf{Triple}} \\ 
\cmidrule(r){2-4} \cmidrule(r){5-7} \cmidrule(r){8-8}
& \textit{RGB} & \textit{IR} & \textit{PC} & \textit{PC+RGB} & \textit{PC+IR} & \textit{RGB+IR} & \textit{RGB+IR+PC} \\ \midrule
Capsule & 0.952 & 0.896 & 0.817 & 0.952 & 0.898 & \textbf{0.977} & 0.967 \\
Cotton & 0.868 & 0.922 & 0.835 & 0.883 & 0.938 & 0.958 & \textbf{0.960} \\
Cube & 0.949 & 0.968 & 0.423 & 0.949 & 0.968 & 0.968 & \textbf{0.980} \\
Spring Pad & 0.871 & 0.779 & 0.817 & \textbf{0.904} & 0.792 & 0.817 & 0.879 \\
Screw & \textbf{1.000} & 0.935 & 0.656 & \textbf{1.000} & 0.948 & 0.974 & \textbf{1.000} \\
Screen & 0.759 & \textbf{0.991} & 0.756 & 0.781 & \textbf{0.991} & 0.981 & 0.938 \\
Piggy & \textbf{1.000} & 0.960 & 0.400 & \textbf{1.000} & 0.960 & \textbf{1.000} & \textbf{1.000} \\
Nut & \textbf{0.976} & 0.590 & 0.541 & 0.955 & 0.590 & 0.769 & 0.959 \\
Flat pad & 0.710 & \textbf{0.950} & 0.830 & 0.780 & \textbf{0.950} & 0.937 & 0.863 \\
Plastic Cylinder & 0.871 & \textbf{1.000} & 0.600 & 0.871 & \textbf{1.000} & \textbf{1.000} & 0.993 \\
Zipper & 0.958 & 0.991 & 0.807 & 0.970 & \textbf{0.994} & 0.988 & \textbf{0.994} \\
Button Cell & \textbf{1.000} & 0.826 & 0.813 & \textbf{1.000} & 0.832 & 0.977 & \textbf{1.000} \\
Toothbrush & 0.951 & 0.920 & 0.920 & \textbf{0.966} & 0.924 & 0.939 & 0.955 \\
Solar Panel & 0.826 & 0.941 & 0.400 & 0.803 & 0.933 & \textbf{0.977} & 0.949 \\
Light & 0.967 & 0.969 & 0.403 & 0.969 & 0.969 & 0.961 & \textbf{0.972} \\
\midrule
Mean & 0.911 & 0.909 & 0.668 & 0.919 & 0.912 & 0.948 & \textbf{0.961}           
\\ \bottomrule   
\end{tabular}}
\label{tab:4}\vspace{-3mm}
\end{table}

\section{Benchmark: MulSen-AD Bench}
\label{Benchmark}

\subsection{Benchmarking Setup}

\textcolor{black}{
\noindent\textbf{Benchmarking method selection.} To thoroughly evaluate sensor data fusion in our multi-sensor anomaly detection setting, we adopt the MulSen-TripleAD algorithm with various sensor combinations as benchmark methods. The corresponding results are shown in Table~\ref{tab:4}. In the table, Single refers to using only RGB, infrared, or point cloud data without decision gating. Double combines two of the three sensor types, while Triple corresponds to the full MulSen-TripleAD pipeline, as illustrated in Figure~\ref{fig:5}. For robust feature extraction, we leverage pretrained PointMAE~\cite{pang2022masked} for point clouds and DINO~\cite{zhang2022dino} for RGB and infrared data. The memory bank follows the Patchcore~\cite{roth2022towards} setup, while the decision gating unit adopts the M3DM~\cite{wang2023multimodal} configuration. All experiments were conducted with identical parameter settings to ensure fair comparisons.
}

\begin{table}[t!]
\centering
\caption{\textcolor{black}{\textbf{Anomaly localization results, measured by pixel-level AUROC ($\uparrow$), pixel-F1-max ($\uparrow$), and pixel-AUPR ($\uparrow$) on the dataset.} `PC' refers to point cloud. `IR' refers to infrared.}}
\vspace{-2mm}
\centering\setlength{\tabcolsep}{2mm}
\resizebox{0.48\textwidth}{!}{
\begin{tabular}{lccccccccc}
\\ \toprule
\multirow{2}{*}{\textbf{Category}} & \multicolumn{3}{c}{\textbf{Pixel-AUROC}} & \multicolumn{3}{c}{\textbf{Pixel-F1-max}} & \multicolumn{3}{c}{\textbf{Pixel-AUPR}} \\ 
\cmidrule(r){2-4} \cmidrule(r){5-7} \cmidrule(r){8-10}
& \textit{RGB} & \textit{IR} & \textit{PC} & \textit{RGB} & \textit{IR} & \textit{PC} & \textit{RGB} & \textit{IR} & \textit{PC} \\ \midrule
Capsule & 0.995 & 0.967 & 0.766 & 0.601 & 0.319 & 0.261 & 0.651 & 0.194 & 0.226 \\
Cotton & 0.999 & 0.954 & 0.641 & 0.773 & 0.361 & 0.141 & 0.829 & 0.334 & 0.070 \\
Cube & 0.989 & 1.000 & 0.754 & 0.495 & 0.777 & 0.154 & 0.471 & 0.840 & 0.064 \\
Spring Pad & 0.997 & 0.946 & 0.742 & 0.299 & 0.410 & 0.297 & 0.198 & 0.289 & 0.222 \\
Screw & 0.991 & 0.996 & 0.527 & 0.318 & 0.290 & 0.059 & 0.275 & 0.184 & 0.028 \\
Screen & 0.855 & 0.930 & 0.539 & 0.332 & 0.170 & 0.067 & 0.259 & 0.104 & 0.039 \\
Piggy & 0.987 & 0.989 & 0.559 & 0.435 & 0.501 & 0.017 & 0.401 & 0.507 & 0.007 \\
Nut & 0.992 & 0.979 & 0.783 & 0.191 & 0.431 & 0.216 & 0.118 & 0.240 & 0.100 \\
Flat Pad & 0.992 & 0.995 & 0.636 & 0.254 & 0.386 & 0.089 & 0.177 & 0.321 & 0.041 \\
Plastic Cylinder & 0.991 & 0.993 & 0.608 & 0.469 & 0.527 & 0.062 & 0.466 & 0.546 & 0.022 \\
Zipper & 0.966 & 0.991 & 0.581 & 0.176 & 0.516 & 0.109 & 0.092 & 0.481 & 0.054 \\
Button Cell & 0.996 & 0.993 & 0.700 & 0.410 & 0.287 & 0.056 & 0.394 & 0.245 & 0.021 \\
Toothbrush & 0.989 & 0.872 & 0.447 & 0.058 & 0.048 & 0.075 & 0.025 & 0.015 & 0.032 \\
Solar Panel & 0.992 & 0.959 & 0.648 & 0.363 & 0.180 & 0.043 & 0.366 & 0.090 & 0.005 \\
Light & 0.998 & 0.993 & 0.543 & 0.395 & 0.458 & 0.013 & 0.335 & 0.381 & 0.006 \\
\midrule
Mean & 0.982 & 0.970 & 0.632 & 0.371 & 0.377 & 0.111 & 0.337 & 0.318 & 0.062 \\
\bottomrule   
\end{tabular}}
\label{tab:5}\vspace{-3mm}
\end{table}

\noindent\textbf{Evaluation metric.} 
Following~\cite{li2023towards,Bergmann_2022}, we use the Area Under the Receiver Operating Characteristic Curve (AUROC) for object-level anomaly detection and employ pixel-level AUROC, pixel-F1-max, and pixel-AUPR for anomaly localization across each modality.


\subsection{Comparisons with SOTA methods on Object-level Anomaly Detection}
Our proposed MulSen-TripleAD model, by leveraging complementary information from three distinct modalities—RGB, infrared, and point cloud—achieves significantly better overall performance than previous SOTA models, which rely on single-modality data. As shown in Table \ref{tab:results}, MulSen-TripleAD outperforms SOTA models by a considerable margin in object-level AUROC across multiple categories, achieving an average AUROC of 0.961. For example, MulSen-TripleAD achieves an AUROC of 1.000 on categories such as Screw, Piggy, and Button Cell, surpassing the closest competing models by at least 3–5\% in these categories. Moreover, our model shows substantial gains in challenging categories like Light and Plastic Cylinder, with improvements of 5.3\% and 3.7\%, respectively. This multi-modal approach enhances anomaly detection by capturing a broader range of feature variations across object surfaces and contexts, enabling our model to detect subtle anomalies that single-modality approaches often miss.

\begin{figure*}[t!]
  \centering
  \setlength{\abovecaptionskip}{0.cm}
  \includegraphics[width=1.0\textwidth]{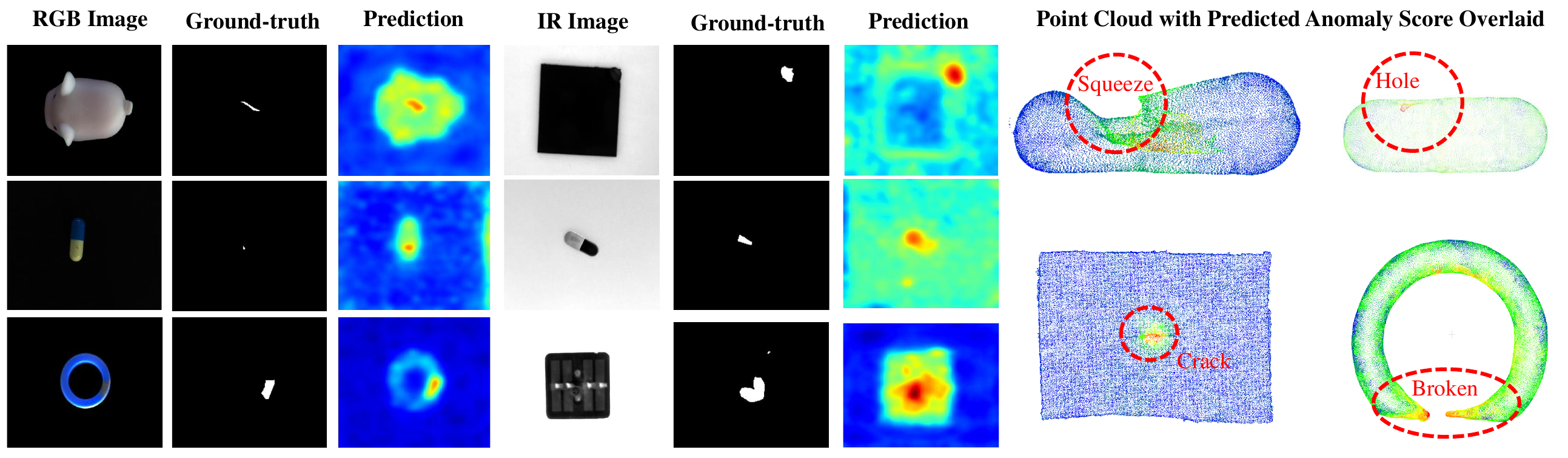}
  \caption{\textcolor{black}{\textbf{Qualitative results of anomaly localization on MulSen-AD dataset.}
}}
  \label{fig:5}\vspace{-2mm}
\end{figure*}

\subsection{ MulSen-TripleAD for Anomaly Detection}

Table~\ref{tab:4} benchmarks various sensor configurations of our MulSen-TripleAD on the MulSen-AD dataset.

\noindent\textbf{Single-sensor performance.}
Individual sensors face clear limitations in detecting the diverse anomalies found in industrial settings. RGB data achieves an AUROC of 91.1\%, performing well for surface-level defects but struggling with subsurface issues. Similarly, infrared imaging, with an AUROC of 90.9\%, excels in subsurface anomalies but falters where geometric accuracy is critical. Point cloud data, despite its utility for 3D geometric features, lags significantly with a 66.8\% AUROC, struggling in categories like `Cube' (42.3\%) and `Solar panel' (40.0\%). These results underscore the insufficiency of single-sensor approaches for comprehensive anomaly detection.

\noindent\textbf{Dual-sensor fusion.}
Dual-sensor combinations yield substantial improvements. RGB + infrared achieves an AUROC of 94.8\%, significantly boosting performance in categories like `Capsule' (97.7\%) and `Piggy' (100\%). Similarly, point cloud + RGB and point cloud + infrared configurations improve to 91.9\% AUROC. However, dual-sensor fusion still leaves gaps, especially where both surface-level and geometric precision are critical, underscoring the need for a more integrated multi-modal approach.

\noindent\textbf{Triple-sensor fusion (MulSen-TripleAD).}
Our MulSen-TripleAD method, combining RGB, infrared, and point cloud data, achieves the highest AUROC of 96.1\%. This comprehensive fusion captures surface, subsurface, and geometric features, offering robust detection in categories like `Cube' (98.0\%), `Nut' (95.9\%), and `Light' (97.2\%). The inclusion of point cloud data proves crucial where 3D geometry matters, while RGB and infrared provide complementary insights, demonstrating the power of multi-sensor integration to address individual modality limitations.

\noindent\textbf{Insights from the MulSen-AD Dataset.}
The MulSen-AD dataset challenges both single- and multi-sensor methods by exposing the complexity of real-world anomalies. Single-sensor performance in categories like `Plastic cylinder' and `Nut' highlights the need for fusion---while infrared detects internal anomalies in `Plastic cylinder' (100\% AUROC), it struggles with geometric accuracy in `Nut', where triple-sensor fusion improves performance to 95.9\%. This validates the dataset’s design in enhancing anomaly detection.



\subsection{\textcolor{black}{{MulSen-TripleAD for Anomaly Localization}}}
We present the anomaly localization results for each modality in Table~\ref{tab:5} and provide following insights:

\noindent\textbf{1) RGB consistently outperforms other modalities in anomaly localization.} RGB achieves the highest average Pixel-AUROC score of 0.982, significantly outperforming Point Cloud (0.632) and slightly surpassing Infrared (0.970). This highlights RGB's strength in capturing detailed visual information necessary for pixel-level anomaly detection across a wide variety of  classes.

\noindent\textbf{2) Infrared shows comparable performance to RGB in precision-recall balancing.} Infrared achieves a Pixel-F1-max score close to RGB (0.377 vs. 0.371), indicating similar effectiveness in precision-recall balancing for anomaly localization. However, each modality brings unique advantages: Infrared excels in detecting temperature and reflectance variations, capturing anomalies that RGB may overlook, while RGB remains more effective for surface-level details. This complementary performance underscores the value of combining modalities to address diverse anomaly characteristics.

\noindent\textbf{3) Point cloud data struggles in fine-grained anomaly localization.} Across all metrics, Point Cloud performs poorly compared to RGB and Infrared, with particularly low scores in Pixel-F1-max (0.111) and Pixel-AUPR (0.062). This indicates significant limitations in its ability to localize anomalies at a detailed, pixel level, especially for complex objects like `zipper' and `spring pad', where capturing intricate geometrical details is critical. These results underscore the need for further refinement of point cloud-based anomaly detection methods to enhance  real-world applicability.

\noindent\textbf{Qualitative results.} 
Fig.~\ref{fig:5} showcases anomaly localization using RGB, infrared, and point cloud data from MulSen-AD dataset. RGB captures surface defects, IR detects subsurface anomalies, and point cloud identifies 3D geometric issues, highlighting the complementary strengths of varied sensors.






\section{Conclusion and Future Work}
\textcolor{black}{In this work, we introduce the first comprehensive framework for Multi-Sensor Anomaly Detection and release the MulSen-AD dataset, specifically designed to evaluate anomaly detection algorithms across multiple sensor modalities. We further propose MulSen-TripleAD, a baseline model that leverages the fusion of RGB, infrared, and point cloud data to address the challenges of unsupervised object-level anomaly detection. Our work opens new avenues for research, encouraging further exploration of sensor fusion techniques and their application in complex, real-world industrial environments.}

\noindent\textbf{Limitation and future work.} While MulSen-AD integrates RGB, infrared, and point cloud data, it lacks deeper-sensing modalities like X-ray , which could enhance detection of internal anomalies. Additionally, our use of decision-level fusion may miss important cross-modal interactions; exploring feature- and modality-level fusion could improve  generalization. The current focus on unsupervised detection leaves room for future exploration of few-shot, zero-shot, and cross-domain settings. Lastly, optimizing scalability and real-time performance remains a challenge, especially for resource-constrained environments, necessitating more efficient fusion methods without sacrificing accuracy.

{
    \small
    \bibliographystyle{ieeenat_fullname}
    \bibliography{main}
}
\clearpage
\clearpage
\setcounter{page}{1}
\maketitlesupplementary

\section{Data Collection Details}
\label{sec:rationale}

As described in Sec 3.3 of the main text, the duration of thermal stimulation in Lock-in infrared thermography depends on the material properties and size of the objects. Objects with higher density and larger volume generally require a longer duration. This duration is controlled by two primary parameters: the lock-in period and the lock-in frequency.  The lock-in period is responsible for the number of thermal stimulation, and the lock-in frequency determines the time intervals between each stimulation. Table 1 provides detailed information on the materials and dimensions of 15 objects, along with their respective lock-in periods and frequencies.

\begin{table}[ht]
\centering
\caption{Properties and Thermal Stimulation Parameters of Objects }
\resizebox{0.5\textwidth}{!}{
\begin{tabular}{clccccccc}
\toprule
& \multirow{2}{*}{\textbf{Category}} 
&  \multicolumn{3}{c}{\textbf{Dimensions [mm]}} 
& \multirow{2}{*}{\textbf{Material}}  
& \multirow{2}{*}{\makecell{\textbf{Lock-in} \\ \textbf{Period}}} 
& \multirow{2}{*}{\makecell{\textbf{Lock-in} \\ \textbf{Frequency [Hz]}}} \\ 
\cmidrule(r){3-5} 
         & &  \textit{Length} & \textit{Width} & \textit{Height} &  & &  \\\midrule
&Capsule & 20.0 & 8.0 & 8.0 & Gelatin           & 30   & 1.0       \\
&Cotton & 80.0 & 80.0 & 1.0  & Fibre          & 30   & 1.0             \\
&Cube & 100.0 & 100.0 &  10.0 & Plastic          & 30   & 0.2         \\
&Piggy & 45.0 &30.0 & 35.0 & Plastic            & 30   & 1.0             \\
&Screen & 130.0 & 60.0 & 0.3 & Glass          & 30   & 1.0         \\
&Flat pad & 16.0 & 16.0 & 0.5 & Metal           & 40   & 1.0           \\
&Screw & 15.0 & 12.0 &  12.0  & Metal            & 60   & 1.0          \\
&Nut & 15.0 & 15.0 & 5.0 & Metal            & 60   & 1.0        \\
&Spring pad & 12.0 & 12.0 &  2.0 & Metal           & 40   & 1.0         \\
&Button cell & 12.0 & 12.0 &  5.0 & Metal           & 50   & 1.0            \\
&Toothbrush & 17.5 & 10.0 &  15.0 & Plastic          & 30   & 2.0           \\
&Zipper & 250.0 & 26.0 &  1.5 & Fibre + Metal         & 30   & 1.0           \\
&Light & 35.0 & 22.0 &  22.0 & Plastic + Metal           & 90   & 0.5           \\
&Plastic cylinder & 30.0 &30.0 &  10.0 & Nylon          & 60   & 1.0           \\
&Solar panel & 40.0 & 40.0 & 3.0 & Silicon           & 30   & 0.2  \\ \bottomrule

\end{tabular}}
\label{tab:data_statistics}
\end{table}

\section{More Dataset Samples}
Due to page limit, only a few dataset samples are shown in the main text. To provide a more intuitive view of our dataset, below are additional dataset samples. Note that the first row represents the RGB image, the second row represents the infrared image, and the third row represents the point cloud. Each column represents a normal or abnormal sample.

\begin{figure}
  \centering
  \setlength{\abovecaptionskip}{0.cm}
  \includegraphics[width=0.5\textwidth]{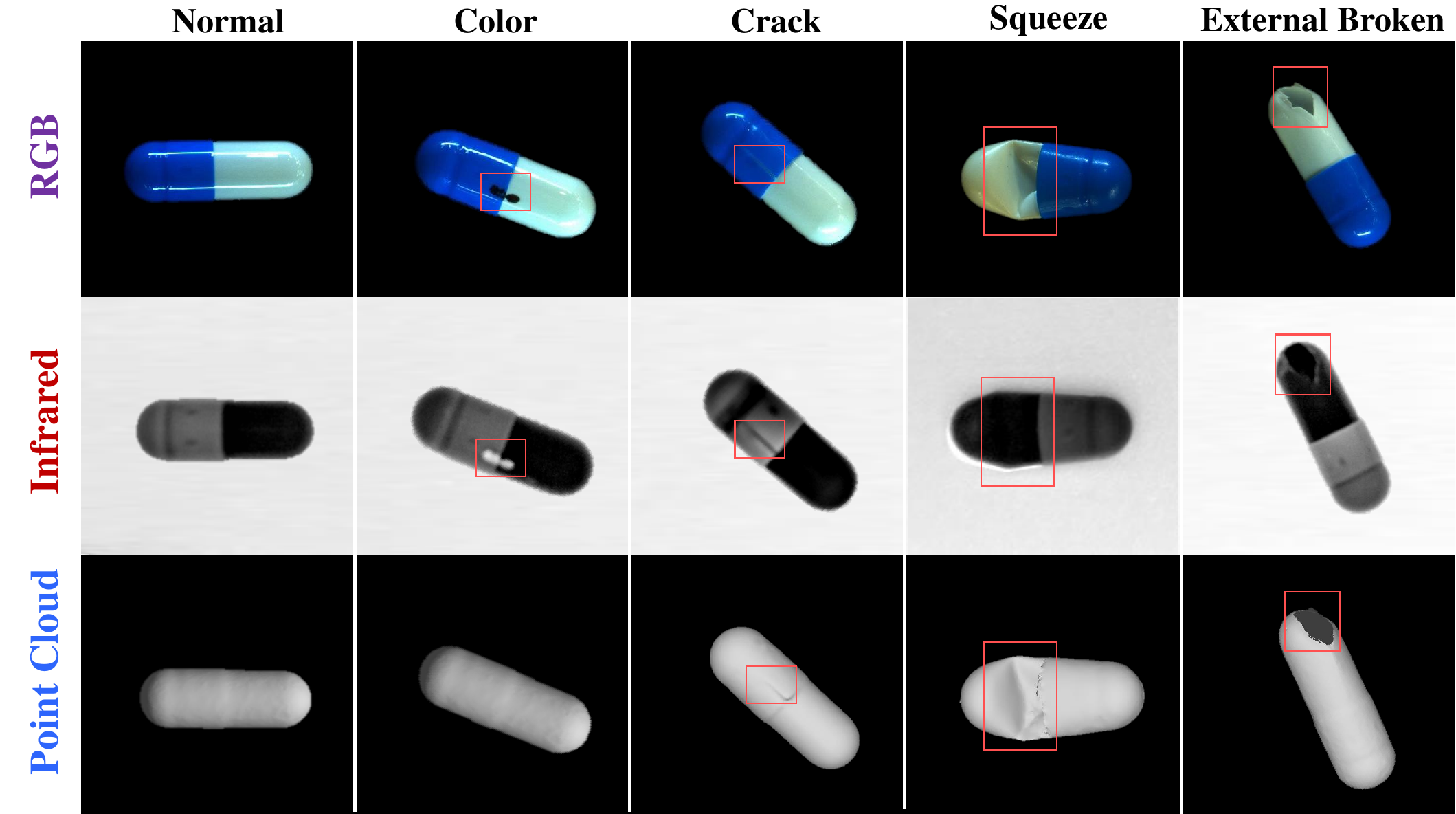}
  \caption{Normal and abnormal \textbf{capsule} samples from the MulSen-AD Dataset. }
  \label{fig:5}
\end{figure}

\begin{figure}
  \centering
  \setlength{\abovecaptionskip}{0.cm}
  \includegraphics[width=0.5\textwidth]{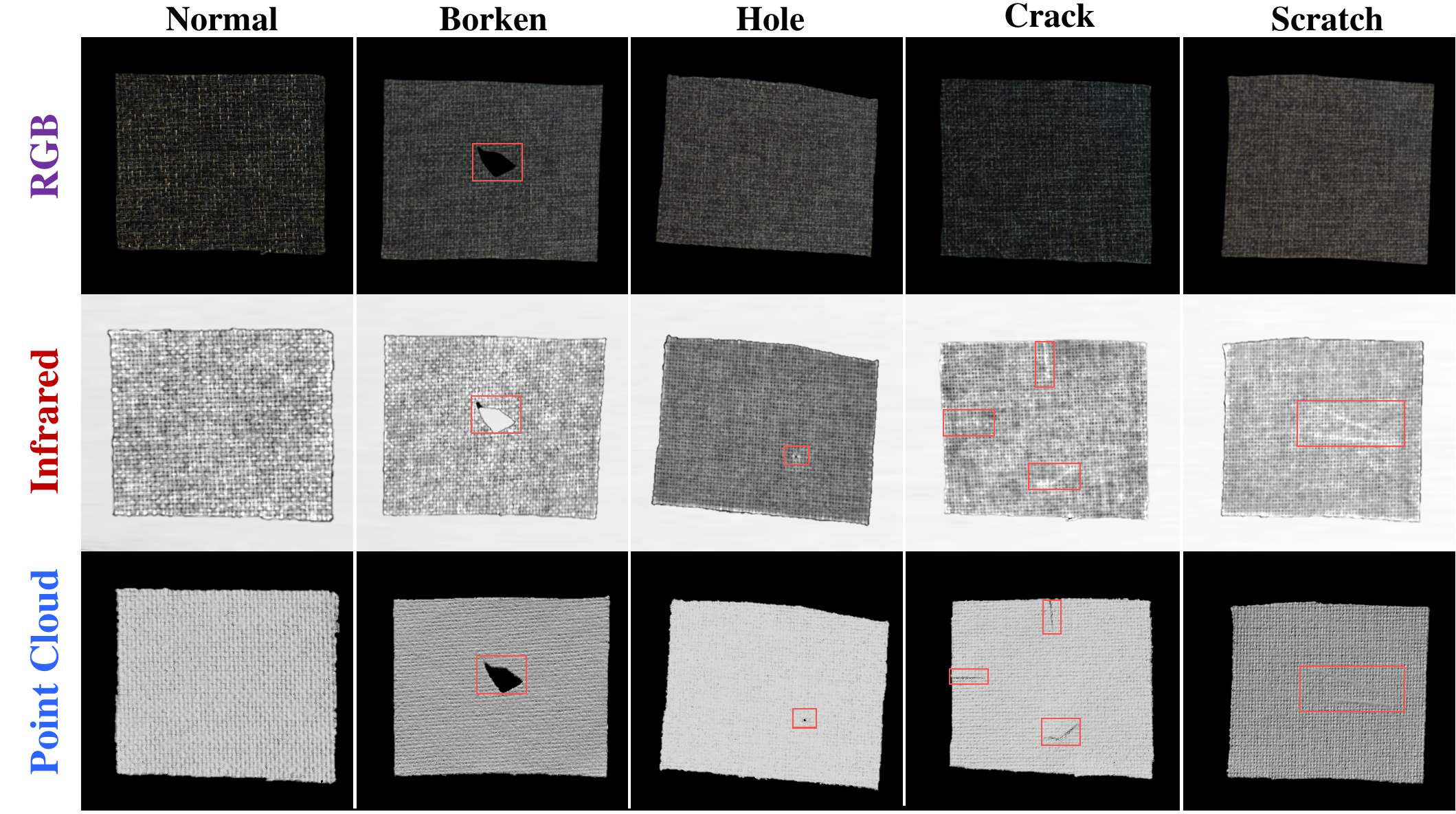}
  \caption{Normal and abnormal \textbf{cotton} samples from the MulSen-AD Dataset.}
  \label{fig:6}
\end{figure}

\begin{figure}
  \centering
  \setlength{\abovecaptionskip}{0.cm}
  \includegraphics[width=0.5\textwidth]{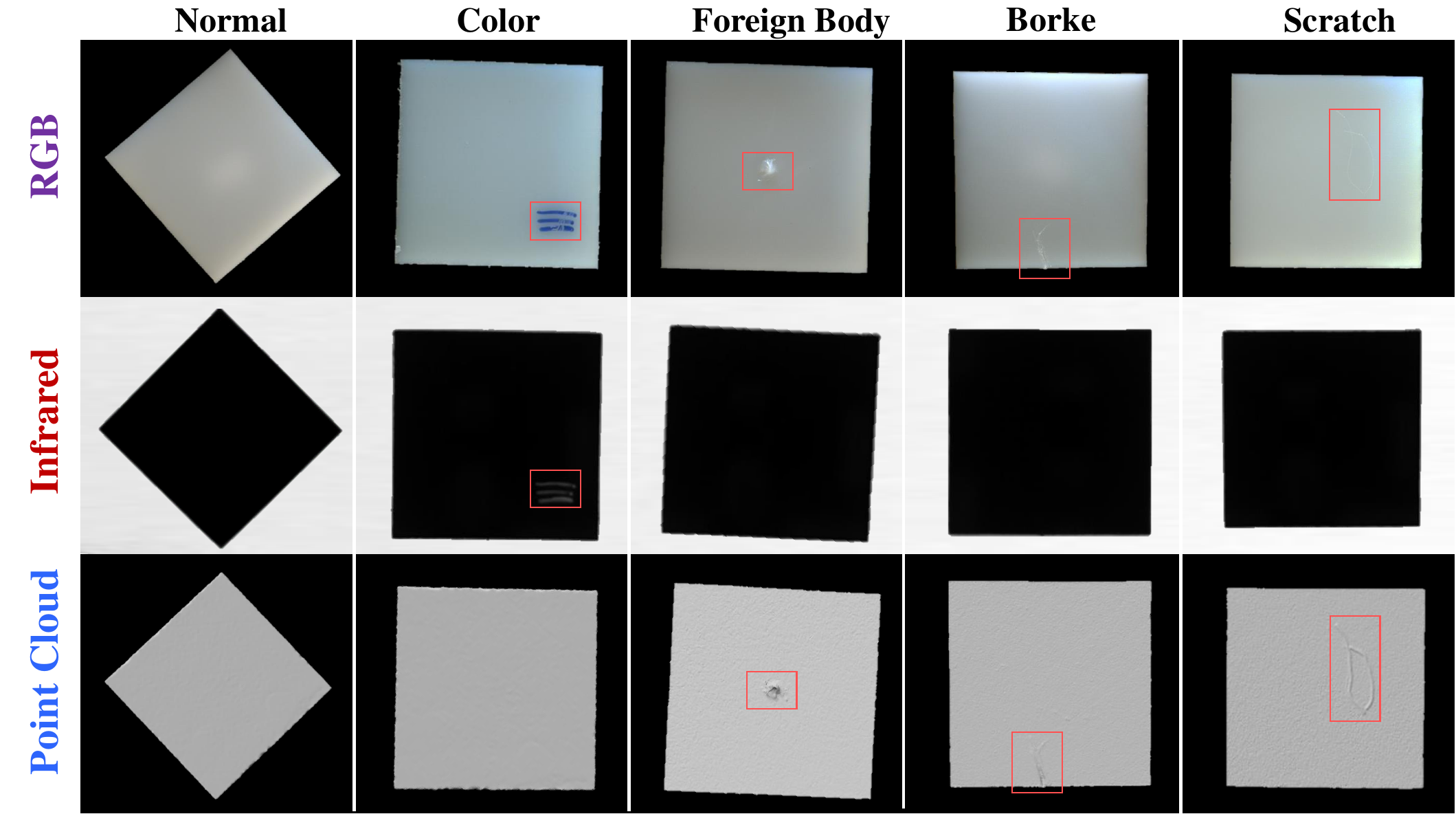}
  \caption{Normal and abnormal \textbf{cube} samples from the MulSen-AD Dataset. }
  \label{fig:7}
\end{figure}

\begin{figure}
  \centering
  \setlength{\abovecaptionskip}{0.cm}
  \includegraphics[width=0.5\textwidth]{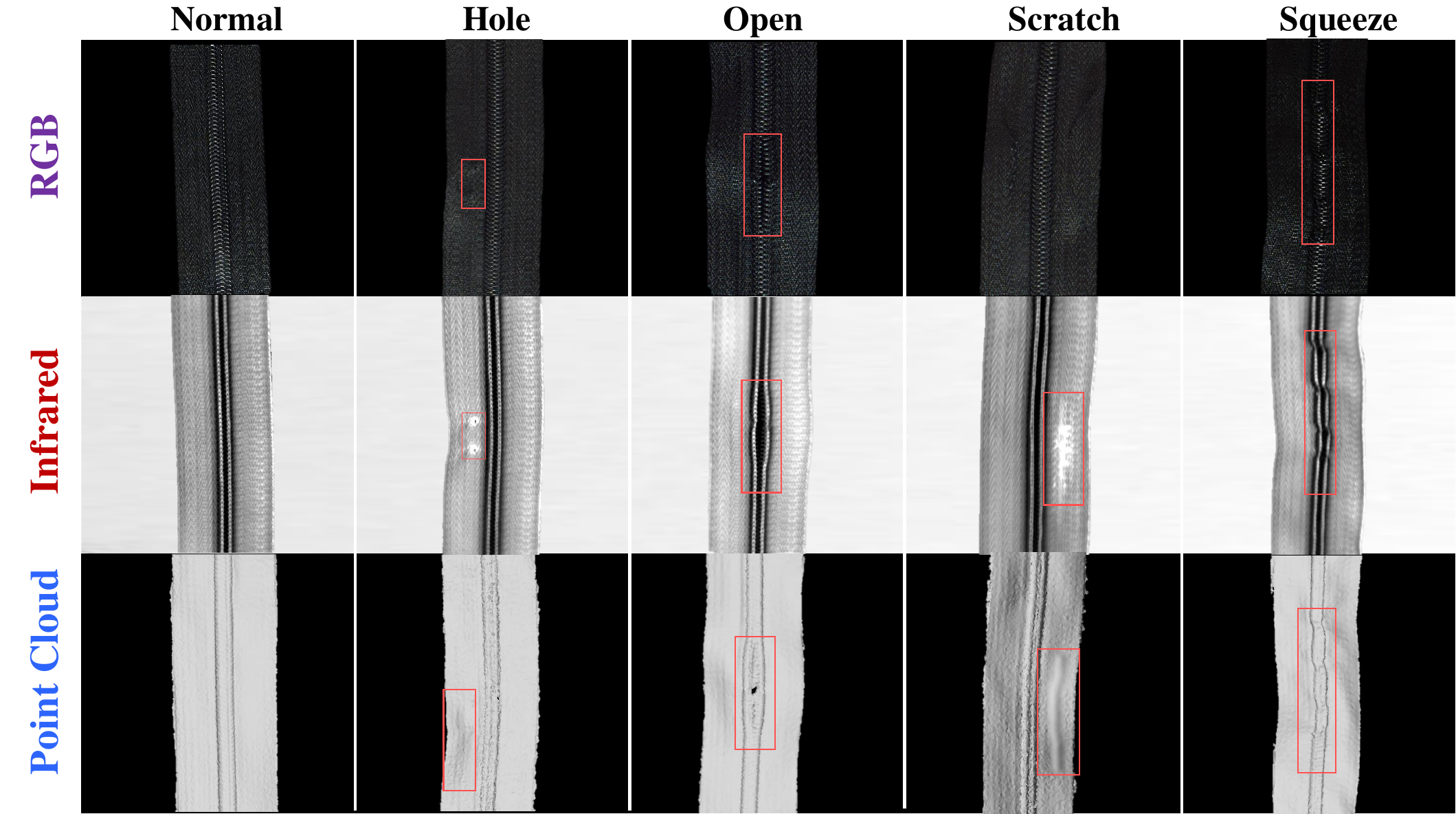}
  \caption{Normal and abnormal \textbf{zipper} samples from the MulSen-AD Dataset.}
  \label{fig:8}
\end{figure}

\begin{figure}
  \centering
  \setlength{\abovecaptionskip}{0.cm}
  \includegraphics[width=0.5\textwidth]{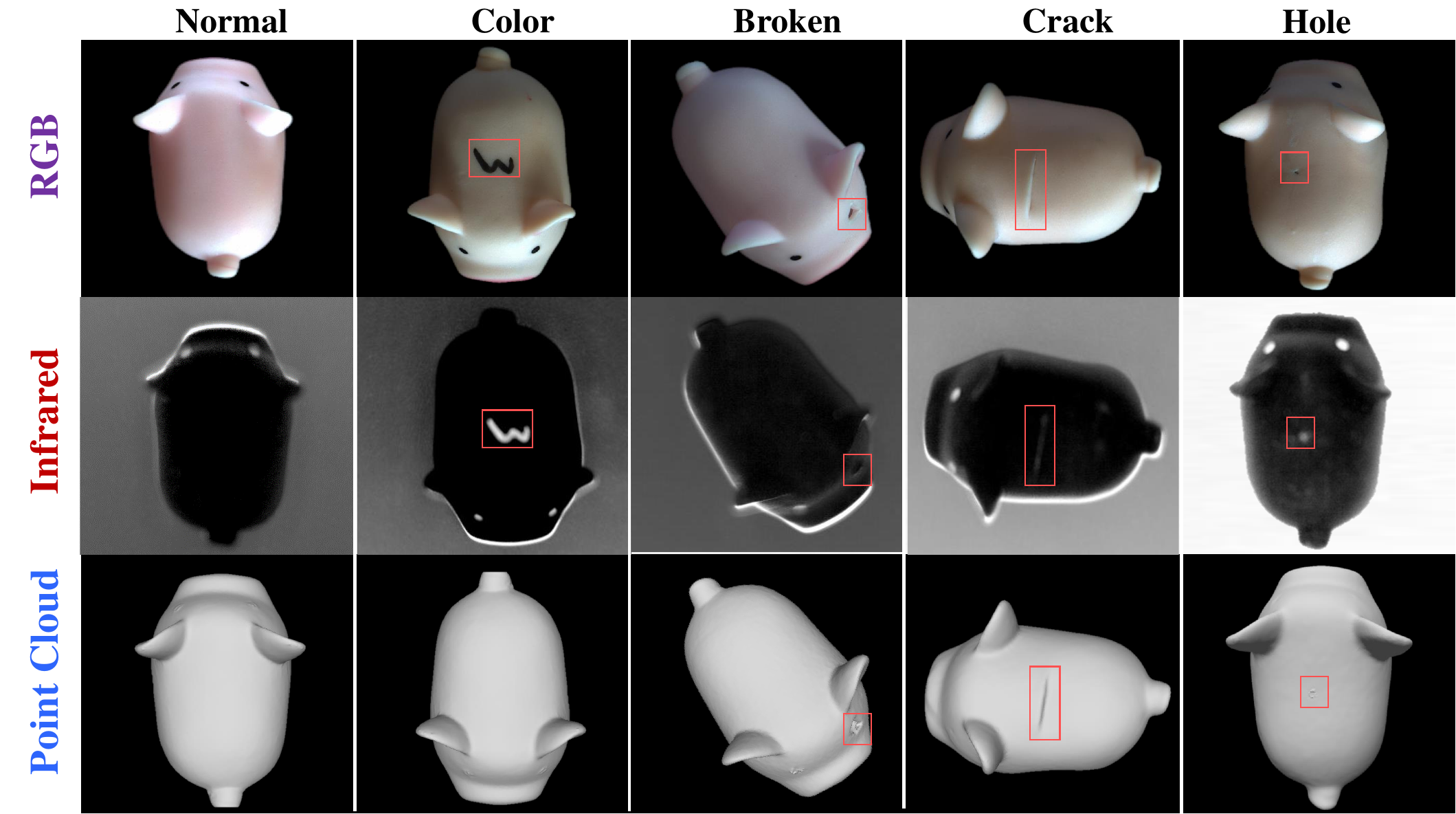}
  \caption{Normal and abnormal \textbf{piggy} samples from the MulSen-AD Dataset.}
  \label{fig:9}
\end{figure}

\begin{figure}
  \centering
  \setlength{\abovecaptionskip}{0.cm}
  \includegraphics[width=0.5\textwidth]{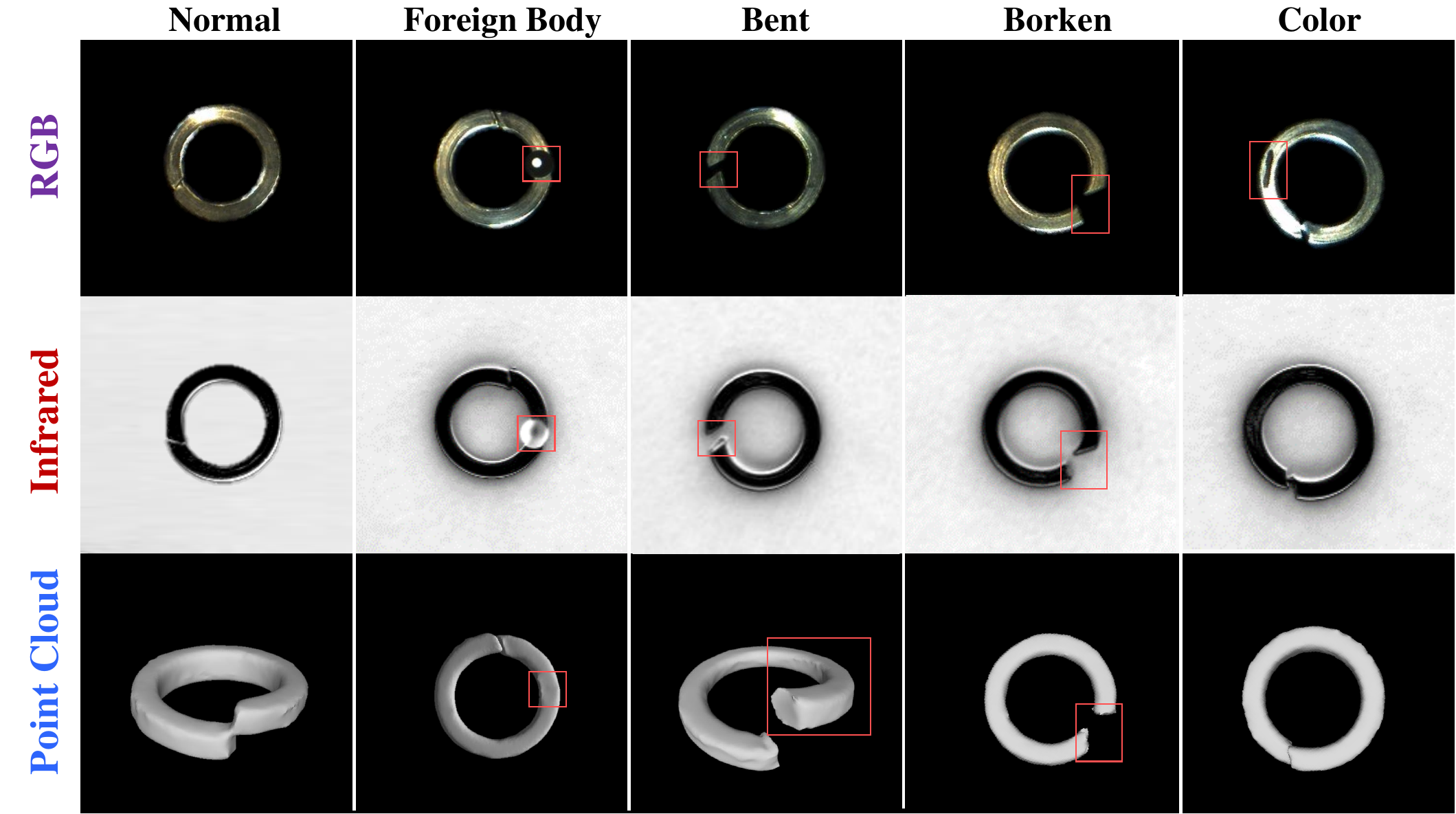}
  \caption{Normal and abnormal \textbf{spring pad} samples from the MulSen-AD Dataset.}
  \label{fig:10}
\end{figure}

\begin{figure}
  \centering
  \setlength{\abovecaptionskip}{0.cm}
  \includegraphics[width=0.5\textwidth]{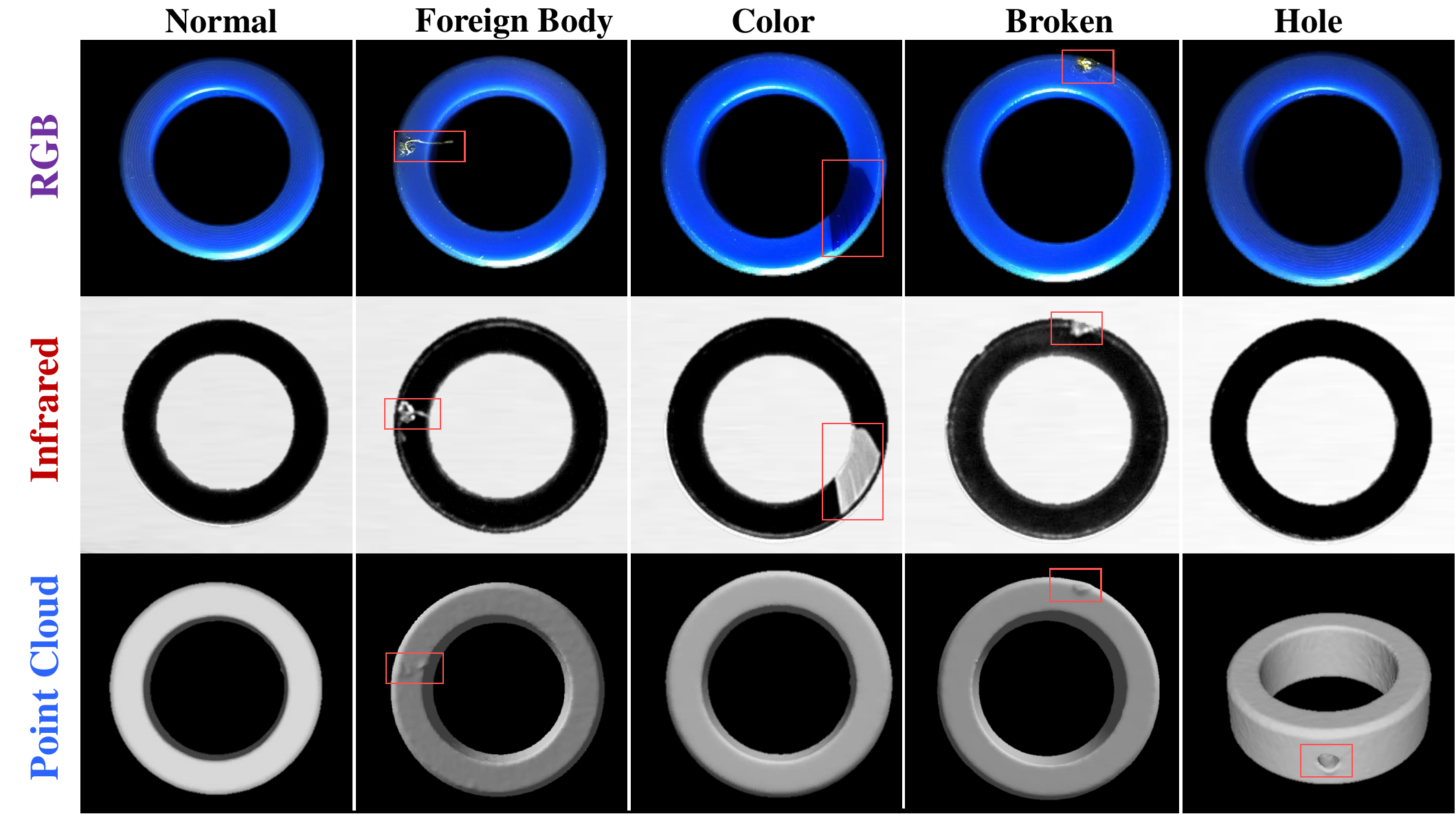}
  \caption{Normal and abnormal \textbf{plastic cylinder} samples from the MulSen-AD Dataset.}
  \label{fig:11}
\end{figure}

\begin{figure}
  \centering
  \setlength{\abovecaptionskip}{0.cm}
  \includegraphics[width=0.5\textwidth]{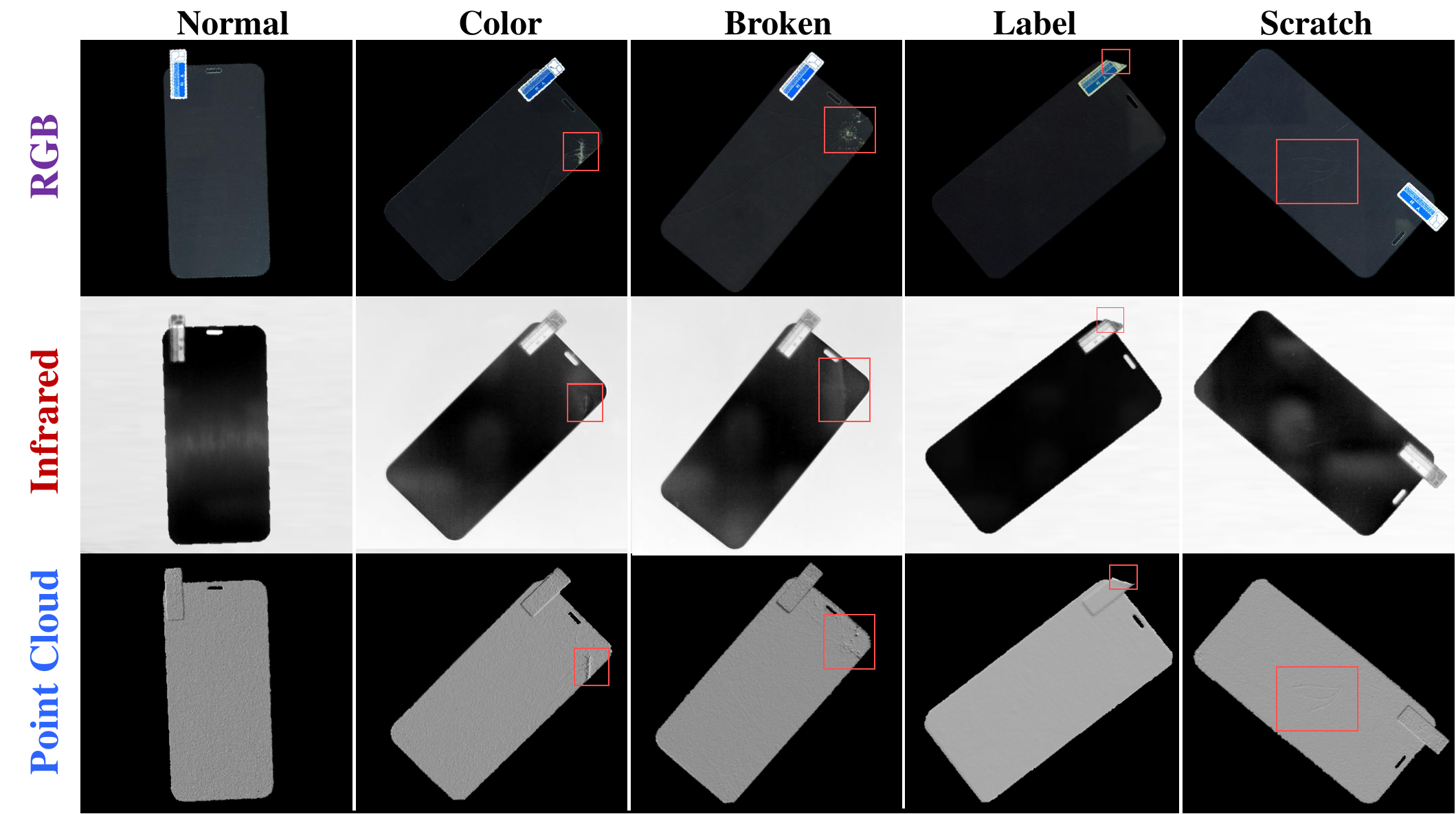}
  \caption{Normal and abnormal \textbf{screen} samples from the MulSen-AD Dataset.}
  \label{fig:12}
\end{figure}

\begin{figure}
  \centering
  \setlength{\abovecaptionskip}{0.cm}
  \includegraphics[width=0.5\textwidth]{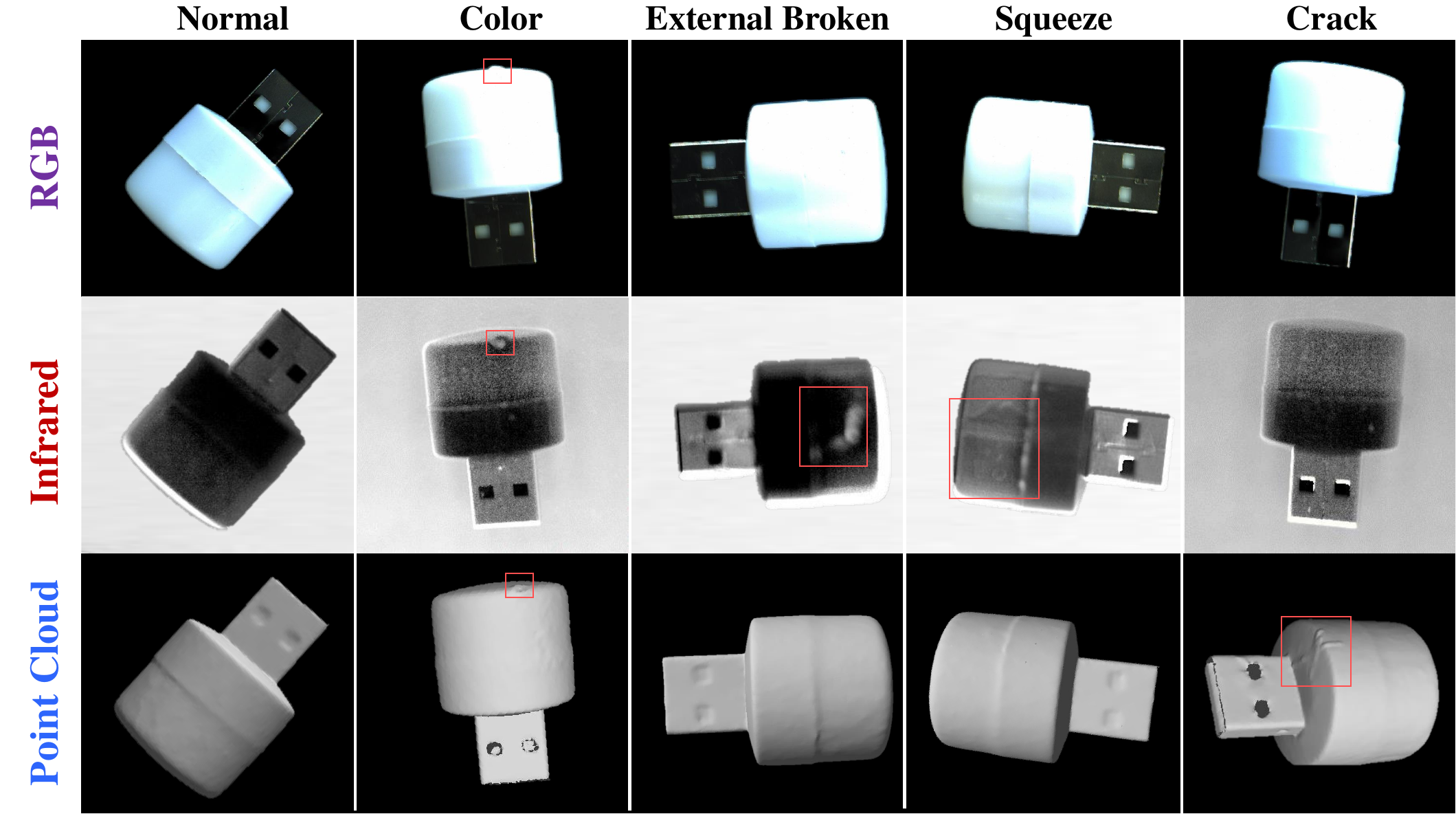}
  \caption{Normal and abnormal \textbf{light} samples from the MulSen-AD Dataset.}
  \label{fig:13}
\end{figure}

\section{Implementation details} \label{sec:experiment_setup}
We employ two Transformer-based feature extractors to independently extract features from RGB/Infrared and point cloud data. For RGB/Infrared feature extraction, we use the ViT-B/8 model, which is pretrained on ImageNet with DINO. This model processes images resized to 224 × 224 pixels and outputs 784 patch features per image. For point cloud feature extraction, we use the PointMAE, pretrained on the ShapeNet dataset. Outputs from layers {3, 7, 11} are used to represent our 3D features. During training, we apply the AdamW optimizer with a learning rate set to 0.001, running the model for 200 epochs. All experiments are conducted on a Tesla V100 GPU.

\section{Single 3D Benchmark}
Due to the page limit, we only give the MulSen-AD Benchmark in the main text. Here we show the Single 3D Benchmark, including object-level Auroc in Table~\ref{tab:3adbench_objauroc}, point-level Auroc in Table~\ref{tab:3adbench_poiauroc}. In the MulSen-AD setting, an object is labeled as abnormal if any one of the three modalities (RGB images, infrared images, or point clouds) is labeled as abnormal. \textbf{However, in the 3D-AD setting, an object is labeled as abnormal only if the point cloud specifically is labeled as abnormal. }

\begin{table*}[htbp]
\centering
\caption{\textsc{SingeBench-3D} for MulSen-AD dataset. The score indicates object-level AUROC $\uparrow$. The best result of each category is highlighted in bold.}
\vspace{-10pt}
\resizebox{0.9\textwidth}{!}{
\begin{tabular}{lccccccccc}
\\ \hline
\multirow{2}{*}{\textbf{Category}} & \multicolumn{2}{c}{\textbf{BTF}} & \multicolumn{2}{c}{\textbf{M3DM}} & \multicolumn{3}{c}{\textbf{PatchCore}}  & \multirow{2}{*}{\textbf{IMRNet}}  & \multirow{2}{*}{\textbf{Reg3D-AD}} \\ 
\cmidrule(r){2-3} \cmidrule(r){4-5} \cmidrule(r){6-8}
 & \textit{Raw} & \textit{FPFH} & \textit{PointMAE} & \textit{PointBERT} & \textit{FPFH} & \textit{FPFH+Raw} & \textit{PointMAE} & \\ \hline
Capsule     & 0.641 & 0.874 & 0.731 & 0.671 & 0.898 & 0.905 & 0.903 &0.601& \textbf{0.912} \\
Cotton      & 0.775 & 0.320 & 0.568 & \textbf{0.805} & 0.253 & 0.263 & 0.197  &0.585& 0.430 \\
Cube        & 0.603 & 0.655 & 0.463 & 0.458 & \textbf{0.723} & 0.668 & 0.722 &0.432& 0.569 \\
Spring pad  & 0.764 & 0.872 & 0.698 & 0.517 & 0.986 & \textbf{1.000} & 0.965 &0.651& 0.951 \\
Screw       & 0.764 & 0.872 & 0.663 & 0.955 & 0.979 & 0.931 & \textbf{0.997} &0.742& 0.972\\
Screen      & 0.584 & 0.788 & 0.906 & 0.928 & 0.916 & \textbf{0.950} & 0.897 &0.378& 0.641 \\
Piggy       & 0.818 & 0.831 & 0.164 & 0.447 & \textbf{1.000} & 0.997 & 0.982 &0.729& 0.866 \\
Nut         & 0.789 & 0.883 & 0.783 & 0.751 & 0.971 & \textbf{0.989} & \textbf{0.989} &0.812& 0.797 \\
Flat pad    & 0.698 & 0.918 & 0.885 & 0.772 & \textbf{1.000} & 0.893 & 0.944 &0.714& 0.908 \\
Plastic cylinder    & 0.728 & 0.866 & 0.462 & 0.706 & \textbf{0.941} & 0.908 & 0.936 &0.621 & 0.765 \\
Zipper      & 0.505 & 0.662 & 0.701 & 0.698 & 0.797 & \textbf{0.813} & 0.739 &0.630& 0.470 \\
Button cell & 0.567 & 0.500 & 0.549 & 0.659 & \textbf{0.915} & 0.687 & 0.797 &0.702& 0.782 \\
Toothbrush  & 0.882 & 0.562 & 0.803 & 0.901 & \textbf{0.905} & 0.888 & 0.891 &0.615& 0.812    \\ 
Solar panel & 0.474 & 0.531 & 0.385 & 0.395 & 0.624 & 0.605 & 0.612 &0.344& \textbf{0.660}    \\
Light       & 0.903 & 0.859 & 0.579 & 0.653 & 0.975 & \textbf{1.000} & 0.992 &0.457& 0.897    \\\hline   
Mean        & 0.711 & 0.721 & 0.628 & 0.705 & \textbf{0.860} & 0.833 & 0.840 &0.601& 0.749    \\ \hline   
\end{tabular}}
\label{tab:3adbench_objauroc}
\end{table*}

\begin{table*}[htbp]
\centering
\caption{\textsc{SingeBench-3D} for MulSen-AD dataset. The score indicates point-level AUROC $\uparrow$. The best result of each category is highlighted in bold.}
\vspace{-10pt}
\resizebox{0.9\textwidth}{!}{
\begin{tabular}{lccccccccc}
\\ \hline
\multirow{2}{*}{\textbf{Category}} & \multicolumn{2}{c}{\textbf{BTF}} & \multicolumn{2}{c}{\textbf{M3DM}} & \multicolumn{3}{c}{\textbf{PatchCore}}  & \multirow{2}{*}{\textbf{IMRNet}} & \multirow{2}{*}{\textbf{Reg3D-AD}} \\ 
\cmidrule(r){2-3} \cmidrule(r){4-5} \cmidrule(r){6-8}
 & \textit{Raw} & \textit{FPFH} & \textit{PointMAE} & \textit{PointBERT} & \textit{FPFH} & \textit{FPFH+Raw} & \textit{PointMAE} & \\ 
\hline
        Capsule & 0.639 & 0.917 & 0.777 & 0.753 & 0.917 & 0.919 & \textbf{0.921} &0.423& 0.877 \\ 
        Cotton & 0.412 & 0.581 & 0.663 & \textbf{0.699} & 0.554 & 0.546 & 0.528 &0.507& 0.521 \\ 
        Cube & 0.441 & \textbf{0.803} & 0.613 & 0.710 & 0.575 & 0.437 & 0.417 &0.566& 0.626 \\ 
        Spring pad  & 0.659 & 0.780 & 0.568 & 0.652 & 0.629 & 0.601 & 0.621 &0.401& \textbf{0.802} \\ 
        Screw & 0.577 & 0.582 & 0.453 & 0.443 & 0.578 & \textbf{0.610} & 0.597 &0.456& 0.540 \\ 
        Screen & 0.469 & \textbf{0.612} & 0.529 & 0.567 & 0.609 & 0.587 & 0.532 &0.352& 0.466 \\ 
        Piggy & 0.735 & \textbf{0.871} & 0.617 & 0.572 & 0.848 & 0.624 & 0.603 &0.512& 0.635 \\ 
        Nut  & 0.640 & \textbf{0.924} & 0.631 & 0.687 & 0.903 & 0.896 & 0.897 &0.369& 0.807 \\ 
        Flat pad & 0.604 & \textbf{0.715} & 0.626 & 0.583 & 0.707 & 0.678 & 0.630 &0.542& 0.692 \\ 
        Plastic cylinder & 0.662 & \textbf{0.858} & 0.510 & 0.652 & 0.830 & 0.766 & 0.769 &0.412& 0.670 \\ 
        Zipper & 0.390 & 0.532 & 0.496 & \textbf{0.563} & 0.552 & 0.545 & 0.502 &0.496& 0.536 \\ 
        Button cell & 0.671 & 0.694 & 0.797 & \textbf{0.799} & 0.382 & 0.512 & 0.478 &0.485& 0.706 \\ 
        Toothbrush & 0.471 & \textbf{0.634} & 0.501 & 0.386 & 0.605 & 0.604 & 0.606 &0.519& 0.472 \\ 
        Solar panel & 0.536 & \textbf{0.727} & 0.539 & 0.601 & 0.202 & 0.265 & 0.274 &0.533& 0.609 \\ 
        Light & 0.665 & \textbf{0.710} & 0.480 & 0.495 & 0.707 & 0.706 & 0.696 &0.415& 0.651 \\ \hline   
        Mean & 0.571 & \textbf{0.729} & 0.587 & 0.611 & 0.640 & 0.620 & 0.605 &0.467& 0.641 \\  \hline   
\end{tabular}}
\label{tab:3adbench_poiauroc}
\end{table*}

\noindent\textbf{Potential negative social impacts.} Our dataset was collected with permission from the factory, so no negative social impact will exist.


\end{document}